\documentclass[11pt,draftcls,onecolumn]{IEEEtran}
\usepackage[cmex10]{amsmath}
\usepackage{epsfig,epsf,psfrag,amssymb,amsfonts,amsthm,latexsym,graphicx,grffile,bm,cite, xcolor,url,subfigure,comment}

\newcommand{\calN}{\mathcal{N}}
\newcommand{\calE}{\mathcal{E}}
\newcommand{\calV}{\mathcal{V}}
\newcommand{\calT}{\mathcal{T}}
\newcommand{\calF}{\mathcal{F}}

\newcommand{\pF}{\widetilde{\mathcal{F}}}
\newcommand{\pT}{\widetilde{\mathcal{T}}}
\newcommand{\calO}{\mathcal{O}}
\newcommand{\calG}{\mathcal{G}}

\newcommand{\bx}{\mathbf{x}}
\newcommand{\bh}{\mathbf{h}}
\newcommand{\bg}{\mathbf{g}}

\newcommand{\bmu}{\boldsymbol{\mu}}

\newcommand{\walksumlong}[3]{\phi(#1\stackrel{#3}{\longrightarrow}#2)}

\newcommand{\deff}{\stackrel{\Delta}{=}}
\newtheorem{proposition}{Proposition}
\newtheorem{theorem}{Theorem}

\usepackage{psfrag}
\author{Ying Liu,~\IEEEmembership{Student Member,~IEEE,} Venkat Chandrasekaran,~\IEEEmembership{Student Member,~IEEE,}\\
        Animashree Anandkumar,~\IEEEmembership{Member,~IEEE,} Alan S. Willsky,~\IEEEmembership{Fellow,~IEEE}
\thanks{\scriptsize This paper was presented in part at the 2010 International Symposium of Information Theory in Austin, Texas, U.S. \cite{Liu&etal:10ISIT}. Y. Liu, V. Chandrasekaran, and A. S. Willsky are with the Stochastic Systems Group, Laboratory for
Information and Decision Systems, Massachusetts Institute of Technology, Cambridge,
MA 02139, USA. Email: \{liu\_ying,venkatc,willsky\}@mit.edu.
A. Anandkumar is with the Center for Pervasive Communications and Computing, University of California, Irvine, CA 92697, USA. Email: a.anandkumar@uci.edu. This research was supported in part by AFOSR through Grant FA9550-08-1-1080 and in part by Shell International Exploration and Production, Inc.}
}
\title{Feedback Message Passing for Inference in Gaussian Graphical Models}

\begin{document}
\maketitle
\vspace{-0.7in}
\begin{abstract}
While loopy belief propagation (LBP) performs reasonably well for inference in some Gaussian graphical models with cycles, its performance is unsatisfactory for many others. In particular for some models LBP does not converge, and in general when it does converge, the computed variances are incorrect (except for cycle-free graphs for which belief propagation (BP) is non-iterative and exact). In this paper we propose {\em feedback message passing} (FMP), a message-passing algorithm that makes use of a special set of vertices (called a {\em feedback vertex set} or {\em FVS}) whose removal results in a cycle-free graph. In FMP, standard BP is employed several times on the cycle-free subgraph excluding the FVS while a special message-passing scheme is used for the nodes in the FVS. The computational complexity of exact inference is $\calO(k^2n)$, where $k$ is the number of feedback nodes, and $n$ is the total number of nodes. When the size of the FVS is very large, FMP is intractable. Hence we propose {\em approximate FMP}, where a pseudo-FVS is used instead of an FVS, and where inference in the non-cycle-free graph obtained by removing the pseudo-FVS is carried out approximately using LBP. We show that, when approximate FMP converges, it yields exact means {\em and} variances on the pseudo-FVS and exact means throughout the remainder of the graph. We also provide theoretical results on the convergence and accuracy of approximate FMP. In particular, we prove error bounds on variance computation. Based on these theoretical results, we design efficient algorithms to select a pseudo-FVS of bounded size. The choice of the pseudo-FVS allows us to explicitly trade off between efficiency and accuracy. Experimental results show that using a pseudo-FVS of size no larger than $\log(n)$, this procedure converges much more often, more quickly, and provides more accurate results than LBP on the entire graph.   
\end{abstract}
\footnote

\IEEEpeerreviewmaketitle
\section{Introduction}
Gaussian graphical models are used to represent the conditional independence relationships among a collection of normally distributed random variables. They are widely used in many fields such as computer vision and image processing \cite{sonka1999image}, gene regulatory networks \cite{werhli2006comparative}, medical diagnostics \cite{heckerman1996causal}, oceanography \cite{wunsch2007practical}, and communication systems \cite{el2001analyzing}. Inference in Gaussian graphical models refers to the problem of estimating the means and variances of all random variables given the model parameters in information form (see Section \ref{sec:graphicalModels} for more details). Exact inference in Gaussian graphical models can be solved by direct matrix inversion for problems of moderate sizes. However, direct matrix inversion is intractable for very large problems involving millions of random variables, especially if variances are sought \cite{fry1993biasing,wunsch2007practical,yang2005ignm}. The development of efficient algorithms for solving such large-scale inference problems is thus of great practical importance. 

Belief propagation (BP) is an efficient message-passing algorithm that gives exact inference results in linear time for tree-structured graphs \cite{jordan2004graphical}. The Kalman filter for linear Gaussian estimation and the forward-backward algorithm for hidden Markov models can be viewed as special instances of BP. Though widely used, tree-structured models (also known as cycle-free graphical models) possess limited modeling capabilities, and many stochastic processes and random fields arising in real-world applications cannot be well-modeled using cycle-free graphs. 

Loopy belief propagation (LBP) is an application of BP on loopy graphs using the same local message update rules. Empirically, it has been observed that LBP performs reasonably well for certain graphs with cycles \cite{murphy1999loopy, crick2003loopy}. Indeed, the decoding method employed for turbo codes has also been shown to be a successful instance of LBP \cite{mceliece1998turbo}. A desirable property of LBP is its distributed nature -- as in BP, message updates in LBP only involve local model parameters or local messages, so all nodes can update their messages in parallel. 

However, the convergence and correctness of LBP are not guaranteed in general, and many researchers have attempted to study the performance of LBP \cite{ihler2006loopy,weiss2001correctness,tatikonda2002loopy,malioutov2006walk}. For Gaussian graphical models, even if LBP converges, it is known that only the means converge to the correct values while the variances obtained are incorrect in general \cite{weiss2001correctness}. In \cite{malioutov2006walk}, a walk-sum analysis framework is proposed to analyze the performance of LBP in Gaussian graphical models. Based on such a walk-sum analysis, other algorithms have been proposed to obtain better inference results \cite{chandrasekaran2008estimation}.

LBP has fundamental limitations when applied to graphs with cycles: Local information cannot capture the global structure of cycles, and thus can lead to convergence problems and inference errors. There are several questions that arise naturally: Can we use more memory to track the paths of messages? Are there some nodes that are more important than other nodes in terms of reducing inference errors? Can we design an algorithm accordingly without losing too much decentralization? 

Motivated by these questions, we consider a particular set of ``important'' nodes called a {\em feedback vertex set} (FVS). A feedback vertex set is a subset of vertices whose removal breaks all the cycles in a graph. In our {\em feedback message passing} (FMP) algorithm, nodes in the FVS use a different message passing scheme than other nodes. More specifically, the algorithm we develop consists of several stages. In the first stage on the cycle-free graph (i.e., that excluding the FVS) we employ standard inference algorithms such as BP but in a non-standard manner: Incorrect estimates for the nodes in the cycle-free portion are computed while other quantities are calculated and then fed back to the FVS. In the second stage, nodes in FVS use these quantities to perform exact mean and variance computations in the FVS and to produce quantities used to initiate the third stage of BP processing on the cycle-free portion in order to correct the means and variances. If the number of feedback nodes is bounded, the means and variances can be obtained exactly in linear time by using FMP. In general, the complexity is $\calO(k^2n)$, where $k$ is the number of the feedback nodes and $n$ is the total number of nodes. 

For graphs with large feedback vertex sets (e.g., for large two-dimensional grids), FMP becomes intractable. We develop {\em approximate FMP} using a pseudo-FVS (i.e., a set of nodes of moderate size that break some but not all of the cycles). The resulting algorithm has the same structure as the exact algorithm except that the inference algorithm on the remainder of the graph, (excluding the pseudo-FVS), which contains cycles, needs to be specified. In this paper we simply use LBP, although any other inference algorithm could also be used. As we will show, assuming convergence of LBP on the remaining graph, the resulting algorithm always yields the correct means {\em and} variances on the pseudo-FVS, and the correct means elsewhere. Using these results and ideas motivated by the work on walk-summability (WS) \cite{malioutov2006walk}, we develop simple rules for selecting nodes for the pseudo-FVS in order to ensure and enhance convergence of LBP in the remaining graph (by ensuring WS in the remaining graph) and high accuracy (by ensuring that our algorithm ``collects the most significant walks''; see Section \ref{sec:walksum} for more details). This pseudo-FVS selection algorithm allows us to trade off efficiency and accuracy in a simple and natural manner. Experimental results suggest that this algorithm performs exceedingly well -- including for non-WS models for which LBP on the entire graph fails catastrophically -- using a pseudo-FVS of size no larger than $\log(n)$.  

Inference algorithms based on dividing the nodes of a graphical model into subsets have been explored previously \cite{pearl1986constraint,darwiche2001recursive}. The approach presented in this paper is distinguished by the fact that our methods can be naturally modified to provide efficient approximate algorithms with theoretical analysis on convergence and error bounds. 

The remainder of the paper is organized as follows. In Section 2, we first introduce some basic concepts in graph theory and Gaussian graphical models. Then we briefly review BP, LBP, and walk-sum analysis. We also define the notion of an FVS and state some relevant results from the literature. In Section 3, we show that for a class of graphs with small FVS, inference problems can be solved efficiently and exactly by FMP. We start with the single feedback node case, and illustrate the algorithm using a concrete example. Then we describe the general algorithm with multiple feedback nodes. We also prove that the algorithm converges and produces correct estimates of the means and variances. In Section 4, we introduce approximate FMP, where we use a pseudo-FVS of bounded size. We also present theoretical results on convergence and accuracy of approximate FMP. Then we provide an algorithm for selecting a good pseudo-FVS. In Section 5, we present numerical results. The experiments are performed on two-dimensional grids, which are widely used in various research areas including image processing. We design a series of experiments to analyze the convergence and accuracy of approximate FMP. We also compare the performance of the algorithm with different choices of pseudo-FVS, and demonstrate that excellent performance can be achieved with a pseudo-FVS of modest size chosen in the manner we describe. Finally in Section 6, we conclude with a discussion of our main contributions and future research directions.

\section{Background}
\label{chp:background}
\subsection{Gaussian Graphical Models}
\label{sec:graphicalModels}

The set of conditional independence relationships among a collection of random variables can be represented by a graphical model \cite{lauritzen1996graphical}. An undirected graph $\calG=(\calV,\calE)$ consists of a set of nodes (or vertices) $\calV$ and a set of edges $\calE$. Each node $s\in\calV$ corresponds to a random variable $x_s$. We say that a set $C\subset \calV$ separates sets $A, B\subset \calV$ if every path connecting $A$ and $B$ passes through $C$. The random vector\footnote{We use the notation $\bx_A$, where $A\subset\calV$, to denote the collection of random variables $\{x_s|s\in A\}$.} $\bx_\calV$ is said to be Markov with respect to $\calG=(\calV,\calE)$ if for any subset $A$, $B$, $C\subset \calV$ where $C$ separates $A$ and $B$, we have that $\bx_A$ and $\bx_B$ are independent conditioned on $\bx_C$, i.e., $p(\bx_A,\bx_B|\bx_C)=p(\bx_A|\bx_C)p(\bx_A|\bx_C)$. Such Markov models on undirected graphs are also commonly referred to as undirected graphical models or Markov random fields.
%

\begin{figure}
\centering
\subfigure[The sparsity pattern of the underlying graph]{
\includegraphics[width=0.38\columnwidth]{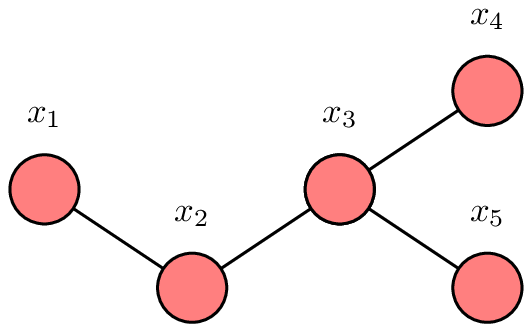}
\label{fig:graphtree}}\hspace{0.5in}
\subfiguretopcapfalse
\subfigure[The sparsity pattern of the information matrix]{
\includegraphics[width=0.38\columnwidth]{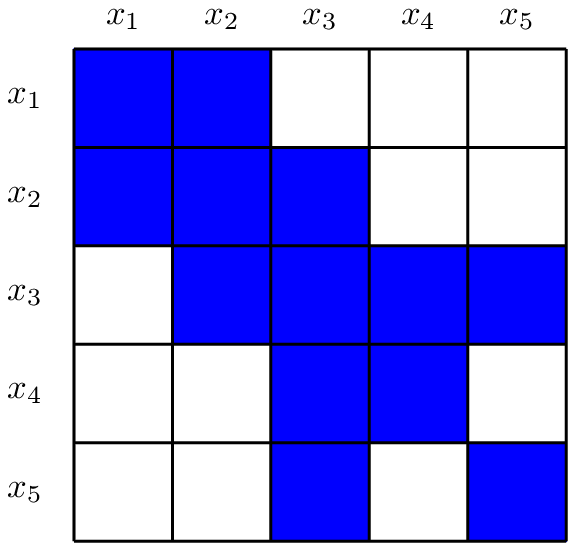}
\label{fig:matrixtree}}
\caption{The relationship between the sparsity pattern in the underlying graph and the sparsity pattern in the information matrix of a Gaussian graphical model. Conditional independences can be directly read from either the sparsity pattern of the graph structure or the sparsity pattern of the information matrix.}
\label{fig:GGM}
\end{figure}

In a Gaussian graphical model, the random vector $x_{\calV}$ is jointly Gaussian. The probability density function of a jointly Gaussian distribution is given by 
$p(\mathbf{x})\propto \exp\{-\frac{1}{2}\mathbf{x}^TJ\mathbf{x}+ \bh^T\mathbf{x}\}$, where $J$ is the {\em information, concentration or precision matrix} and $\bh$ is the {\em potential vector}. We refer to these parameters as the model parameters in information form. The mean vector $\bmu$ and covariance matrix $P$ are related to $J$ and $\bh$ by $\bmu=J^{-1}\bh$ and $P=J^{-1}$. For Gaussian graphical models, the graph structure is sparse with respect to the information matrix $J$, i.e., $J_{i,j} \ne 0$ if and only if there is an edge between $i$ and $j$. For example, Figure \ref{fig:graphtree} is the underlying graph for the information matrix $J$ with sparsity pattern shown in Figure \ref{fig:matrixtree}. For a non-degenerate Gaussian distribution, $J$ is positive definite. The conditional independences of a collection of Gaussian random variables can be read immediately from the graph as well as from the sparsity pattern of the information matrix. If $J_{ij}=0, i\neq j$, then $x_i$ and $x_j$ are independent conditioned on all other variables \cite{speed1986gaussian}. Inference in Gaussian graphical models refers to the problem of estimating the means $\mu_i$ and variances $P_{ii}$ of every random variable $x_i$ given $J$ and $\bh$.

\subsection{Belief Propagation and Loopy Belief Propagation}

BP is a message passing algorithm for solving inference problems in graphical models. Messages are updated at each node according to incoming messages from neighboring nodes and local parameters. It is known that for tree-structured graphical models, BP runs in linear time (in the cardinality $n=|\calV|$ of the node set) and is exact. When there are cycles in the graph, LBP is used instead, where the same local message update rules as BP are used neglecting the existence of cycles. However, convergence and correctness are not guaranteed when there are cycles. 

In Gaussian graphical models, the set of messages can be represented by $\{\Delta J_{i\rightarrow j}\cup \Delta h_{i\rightarrow j}\}_{(i,j)\in\calE}$. Consider a Gaussian graphical model: $p(\bx)\propto \exp\{-\frac{1}{2}\bx^TJ\bx+\bh^T\bx\}$. BP (or LBP) proceeds as follows \cite{malioutov2006walk}:

\begin{enumerate}
\item[(1)]{ Message Passing:}

The messages are initialized as $\Delta J^{(0)}_{i\rightarrow j}$ and $\Delta h^{(0)}_{i\rightarrow j}$, for all $(i,j)\in \calE$. These initializations may be chosen in different ways. In our experiments we initialize all messages with the value $0$.

At each iteration $t$, the messages are updated based on previous messages as
\begin{align}
\Delta J^{(t)}_{i \rightarrow j}&=-J_{ji}(\hat{J}^{(t-1)}_{i\backslash j})^{-1}J_{ij},\\
\Delta h^{(t)}_{i\rightarrow j}&=-J_{ji}(\hat{J}^{(t-1)}_{i\backslash j})^{-1}\hat{h}^{(t-1)}_{i\backslash j},
\label{eq:GBPmessage}
\end{align}
\noindent where 
\begin{align}
\hat{J}^{(t-1)}_{i\backslash j}&=J_{ii}+\sum_{k\in \mathcal{N}(i)\backslash j}{\Delta J^{(t-1)}_{k\rightarrow i}},\\
\hat{h}^{(t-1)}_{i\backslash j}&=h_i+\sum_{k\in \mathcal{N}(i)\backslash j}{\Delta h^{(t-1)}_{k\rightarrow i}.}
\end{align}

Here $\calN(i)= \{j \in \calV: (i,j)\in \calE\}$ denotes the set of neighbors of node $i$. The fixed-point messages are denoted as $\Delta J_{i\rightarrow j}$ and $\Delta h_{i\rightarrow j}$ if the messages converge. 

\item[(2)]{ Computation of Means and Variances:}

The variances and means are computed based on the fixed-point messages as
\begin{align}
\hat{J}_{i}&=J_{ii}+\sum_{k\in \calN(i)}{\Delta J_{k\rightarrow i}},\\
\hat{h}_i&=h_i+\sum_{k\in \calN(i)}{\Delta h_{k\rightarrow i}}.
\label{eq:GBPmarginal}
\end{align}

The variances and means can then be obtained by $P_{ii}=\hat{J}^{-1}_i$ and $\mu_i=\hat{J}^{-1}_{i}\hat{h}_i$.

\end{enumerate}
\subsection{Walk-sum Analysis}
\label{sec:walksum}
Computing means and variances for a Gaussian graphical model corresponds to solving a set of linear equations and obtaining the diagonal elements of the inverse of $J$ respectively. There are many ways in which to do this -- e.g., by direct solution, or using various iterative methods. As we outline in this section, one way to interpret the exact or approximate solution of this problem is through walk-sum analysis, which is based on a simple power series expansion of $J^{-1}$. In \cite{malioutov2006walk,chandrasekaran2008estimation} walk-sum analysis is used to interpret the computations of means and variances formally as collecting all required ``walks'' in a graph. The analysis in \cite{malioutov2006walk} identifies when LBP fails, in particular when the required walks cannot be summed in arbitrary orders, i.e., when the model is not walk-summable.\footnote{Walk-summability corresponds to the absolute convergence of the series corresponding to the walk-sums needed for variance computation in a graphical model \cite{malioutov2006walk}.} One of the important benefits of walk-sum analysis is that it allows us to understand what various algorithms compute and relate them to the required exact computations. For example, as shown in \cite{malioutov2006walk}, LBP collects all of the required walks for the computation of the means (and, hence, if it converges always yields the correct means) but only some of the walks required for variance computations for loopy graphs (so, if it converges, its variance calculations are not correct).

For simplicity, in the rest of the paper, we assume without loss of generality that the information matrix $J$ has been normalized such that all its diagonal elements are equal to unity. Let $R=I-J$, and note that $R$ has zero diagonal. The matrix $R$ is called the {\em edge-weight matrix}.\footnote{The matrix   $R$, which has the same off-diagonal sparsity pattern as $J$, is a matrix of partial correlation coefficients: $R_{ij}$ is the conditional correlation coefficient between $x_i$ and $x_j$ conditioned on all of the other variables in the graph.}

A walk of length $l\geq 0$ is defined as a sequence of vertices $w=(w_0,w_1,w_2,...,w_l)$ where each step $(w_i,w_{i+1})$ is an edge in the graph. The {\em weight of a walk} is defined as the product of the edge weights,
\begin{equation}
\phi(w)=\prod_{l=1}^{l(w)}{R_{w_{l-1},w_{l}}},
\end{equation}
\noindent where $l(w)$ is the length of walk $w$. Also, we define the weight of a zero-length walk, i.e., a single node, as one.

By the Neumann power series for matrix inversion, the covariance matrix can be expressed as 
\begin{equation}
P=J^{-1}=(I-R)^{-1}=\sum_{l=0}^{\infty}{R^l}.
\end{equation}
\noindent This formal series converges (although not necessarily absolutely) if the spectral radius, $\rho(R)$, i.e., the magnitude of the largest eigenvalue of $R$, is less than $1$.

Let $\mathcal{W}$ be a set of walks. We define the walk-sum of $\mathcal{W}$ as
\begin{equation}
\phi(\mathcal{W})\stackrel{\Delta}{=}\sum_{w\in \mathcal{W}} {\phi(w)}.
\end{equation}
We use $\phi(i\rightarrow j)$ to denote the sum of all walks from node $i$ to node $j$. In particular, we call $\phi(i\rightarrow i)$ the {\em self-return walk-sum} of node $i$. 

It is easily checked that the $(i,j)$ entry of $R^l$ equals $\phi^l (i\to j)$, the sum of all walks of length $l$ from node $i$ to node $j$. Hence
\begin{equation}
P_{ij}=\phi(i\rightarrow j)=\sum_{l=0}^{\infty}{\phi^l(i\stackrel{}{\rightarrow}j)}.
\label{eq:summand}
\end{equation}

A Gaussian graphical model is {\em walk-summable} (WS) if for all $i,j\in \mathcal{V}$, the walk-sum $\phi(i\rightarrow j)$ converges for any order of the summands in (\ref{eq:summand}) (note that the summation in (\ref{eq:summand}) is ordered by walk-length).

In walk-summable models, $\phi(i\rightarrow j)$ is well-defined for all $i,j\in\calV$. The covariances and the means can be expressed as
\begin{align}
P_{ij}&=\phi(i\rightarrow j),\hspace{12pt}\\
\mu_{i}&=\sum_{j\in \mathcal{V}}{h_{j} P_{ij}}=\sum_{j\in \mathcal{V}}{h_j \phi(i\rightarrow j)}.
\label{eq:inference}
\end{align}

As shown in \cite{malioutov2006walk} for non-WS models, LBP may not converge and can, in fact, yield oscillatory variance estimates that take on negative values.

Here we list some useful results from \cite{malioutov2006walk} that will be used in this paper.

\begin{enumerate}
\item
The following conditions are equivalent to walk-summability:

(i) $\sum_{w\in \mathcal{W}_{i\to j}}|\phi(w)|$ converges for all $i, j \in \calV$, where $\mathcal{W}_{i\to j}$ is the set of walks from $i$ to $j$.

(ii) $\rho(\bar{R})<1$, where $\bar{R}$ is the matrix whose elements are the absolute values of the corresponding elements in $R$. 
\label{prop:walk-summable}

\item
A Gaussian graphical model is walk-summable if it is attractive, i.e., every edge weight $R_{ij}$ is nonnegative. The model is also walk-summable if the graph is cycle-free.

\item
For a walk-summable Gaussian graphical model, LBP converges and gives the correct means.
\label{prop:LBPconverge}
\item
In walk-summable models, the estimated variance from LBP for a node is the sum over all backtracking walks\footnote{A backtracking walk of a node is a self-return walk that can be reduced consecutively to a single node. Each reduction is to replace a subwalk of the form $\{i,j,i\}$ by the single node $\{i\}$. For example, a self-return walk of the form $12321$ is backtracking, but a walk of the form $1231$ is not.}, which is a subset of all self-return walks needed for computing the correct variance.
\label{prop:backtracking}
\end{enumerate}
\subsection{Feedback Vertex Set}
\label{subsec:FVS}

\begin{figure}
\centering
\subfigure[A graph with an FVS of size one]{
\label{fig:FVS11}
\psfrag{?1?}[c][c][1]{$1$}
\includegraphics[width=0.3\columnwidth]{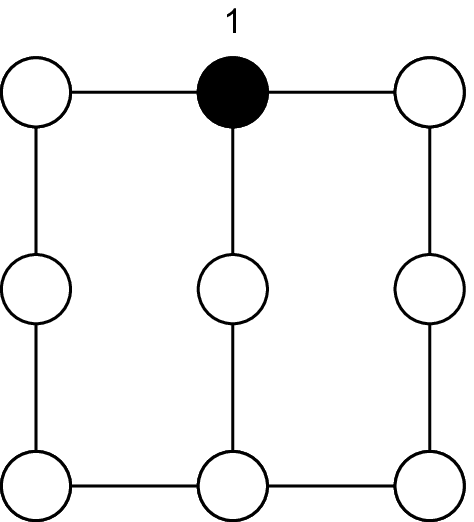}
}\hspace{10pt}
\subfigure[A graph with an FVS of size two]{
\psfrag{1}[c][c][1]{$1$}
\psfrag{2}[c][c][1]{$2$}
\includegraphics[width=0.45\columnwidth]{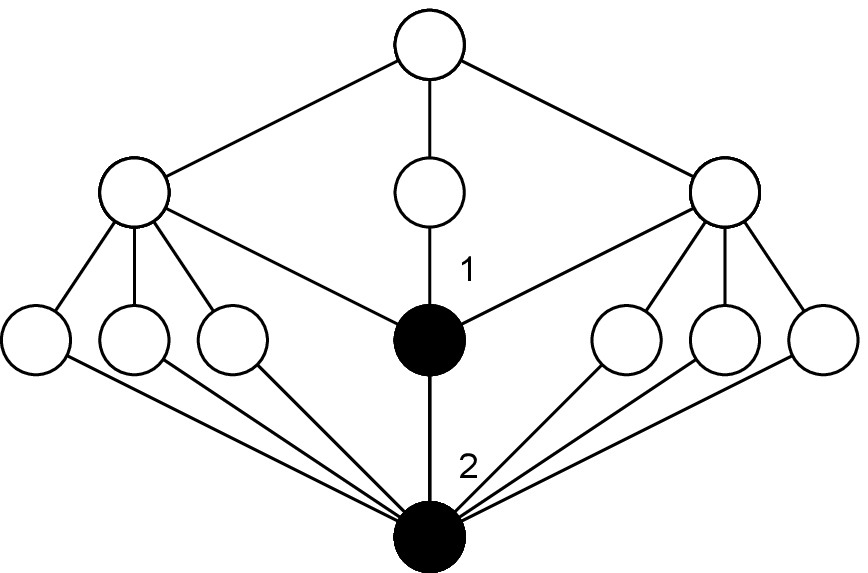}
\label{fig:FVS2}
}
\caption{Examples of FVS's of different sizes. After removing the nodes in an FVS and their incident edges, the reminder of the graph is cycle-free.}
\label{fig:FVS}
\end{figure}

A {\em feedback vertex set} (FVS), also called a loop cutset, is defined as a set of vertices whose removal (with the removal of incident edges) results in an cycle-free graph \cite{vazirani2004approximation}. For example, in Figure \ref{fig:FVS11}, node $1$ forms an FVS by itself since it breaks all cycles. In Figure \ref{fig:FVS2}, the set consisting of nodes $1$ and $2$ is an FVS.
The problem of finding the FVS of the minimum size is called the {\em minimum feedback vertex set} problem, which has been widely studied in graph theory and computer science. For a general graph, the decision version of the minimum FVS problem, i.e., deciding whether there exists an FVS of size at most $k$, has been proved to be NP-complete \cite{karp1972reducibility}. Finding the minimum FVS for general graphs is still an active research area. To the best of the authors' knowledge, the fastest algorithm for finding the minimum FVS runs in time $\calO(1.7548^n)$, where $n$ is the number of nodes \cite{fomin2008minimum}. 

Despite the difficulty of obtaining the minimal FVS, approximate algorithms have been proposed to give an FVS whose size is bounded by a factor times the minimum possible size\cite{erdos1962maximal,bar1994approximation,bafna1992}. In \cite{bafna1992}, the authors proposed an algorithm that gives an FVS of size at most two times the minimum size. The complexity of this algorithm is $\calO(\min\{m\log n,n^2\})$, where $m$ and $n$ are respectively the number of edges and vertices. In addition, if one is given prior knowledge of the graph structure, optimal or near optimal solutions can be found efficiently or even in linear time for many special graph structures \cite{shamir1979linear, wang1985feedback,kratsch2008feedback}. Fixed-parameter polynomial-time algorithms are also developed to find the minimum FVS if the minimum size is known to be bounded by a parameter \cite{dehne20072}.

\section{Exact Feedback Message Passing}
\label{chap:exact FMP}

In this section, we describe the exact FMP algorithm (or simply FMP) which gives the exact inference results for all nodes. We initialize FMP by selecting an FVS, $\calF$, using any one of the algorithms mentioned in Section \ref{subsec:FVS}. The nodes in the FVS are called feedback nodes.

We use a special message update scheme for the feedback nodes while using standard BP messages (although, as we will see, not in a standard way) for the non-feedback nodes. In FMP, two rounds of BP message passing are performed with different parameters. In the first round of BP, we obtain inaccurate ``partial variances'' and ``partial means'' for the nodes in the cycle-free graph as well as some ``feedback gains'' for the non-feedback nodes. Next we compute the exact inference results for the feedback nodes. In the second round of standard BP, we make corrections to the ``partial variances'' and ``partial means'' of the non-feedback nodes. Exact inference results are then obtained for all nodes.

Before describing FMP, we introduce some notation. With a particular choice, $\calF$, of FVS and with $\calT = \calV\slash \calF$ as the remaining cycle-free graph, we can define submatrices and subvectors respectively of $J$ and $\bh$.  In particular, let $J_\calF$ denote the information matrix restricted to nodes of $\calF$ -- i.e., for convenience we assume we have ordered the nodes in the graph so that $\calF$ consists of the first $k$ nodes in $\calV$, so that $J_\calF$ corresponds to the upper-left $k\times k$ block of $J$, and similarly $J_\calT$, the information matrix restricted to nodes in $\calT$ corresponds to the lower right $(n-k)\times (n-k)$ block of $J$. We can also define $J_{\calT\calF}$, the lower left cross-information matrix, and its transpose (the upper-right cross-information matrix) $J_{\calF\calT}$. Analogously we can define the subvectors $\bh_{\calF}$ and $\bh_{\calT}$. In addition, for the graph $\calG$ and any node $j$, let $\calN(j)$ denote the neighbors of $j$, i.e., the nodes connected to $j$ by edges.

In this section we first describe FMP for the example in Figure \ref{fig:1-1}, in which the FVS consists of a single node. Then we describe the general FMP algorithm with multiple feedback nodes. We also prove the correctness and analyze the complexity.

\subsection{The Single Feedback Node Case}

\begin{figure}
\centering
\subfigure[A graph with cycles]{
\label{fig:1-1}
\includegraphics[width=0.3\columnwidth]{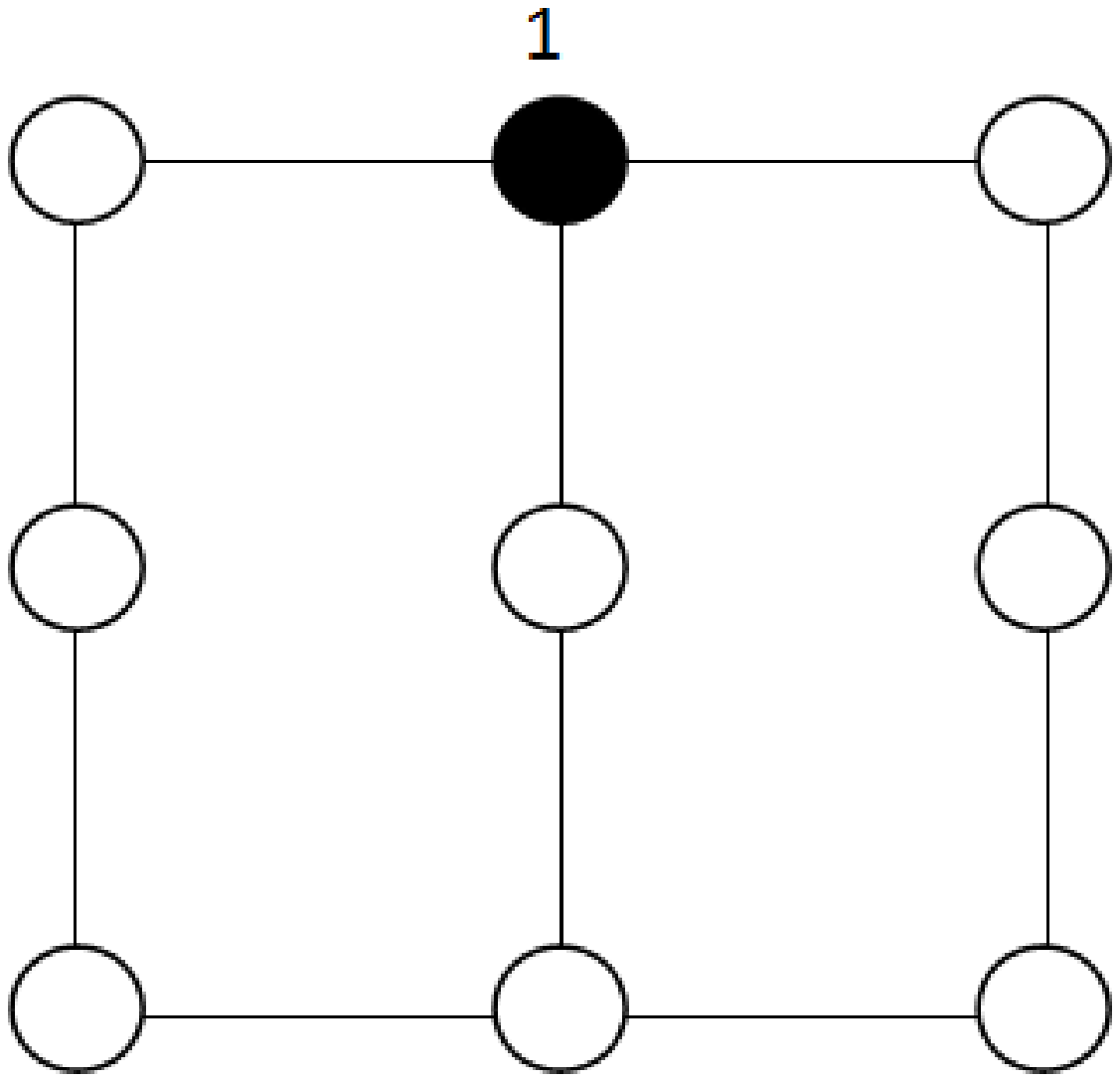}}
\subfigure[Message initialization]{
\label{fig:1-2}
\includegraphics[width=0.3\columnwidth]{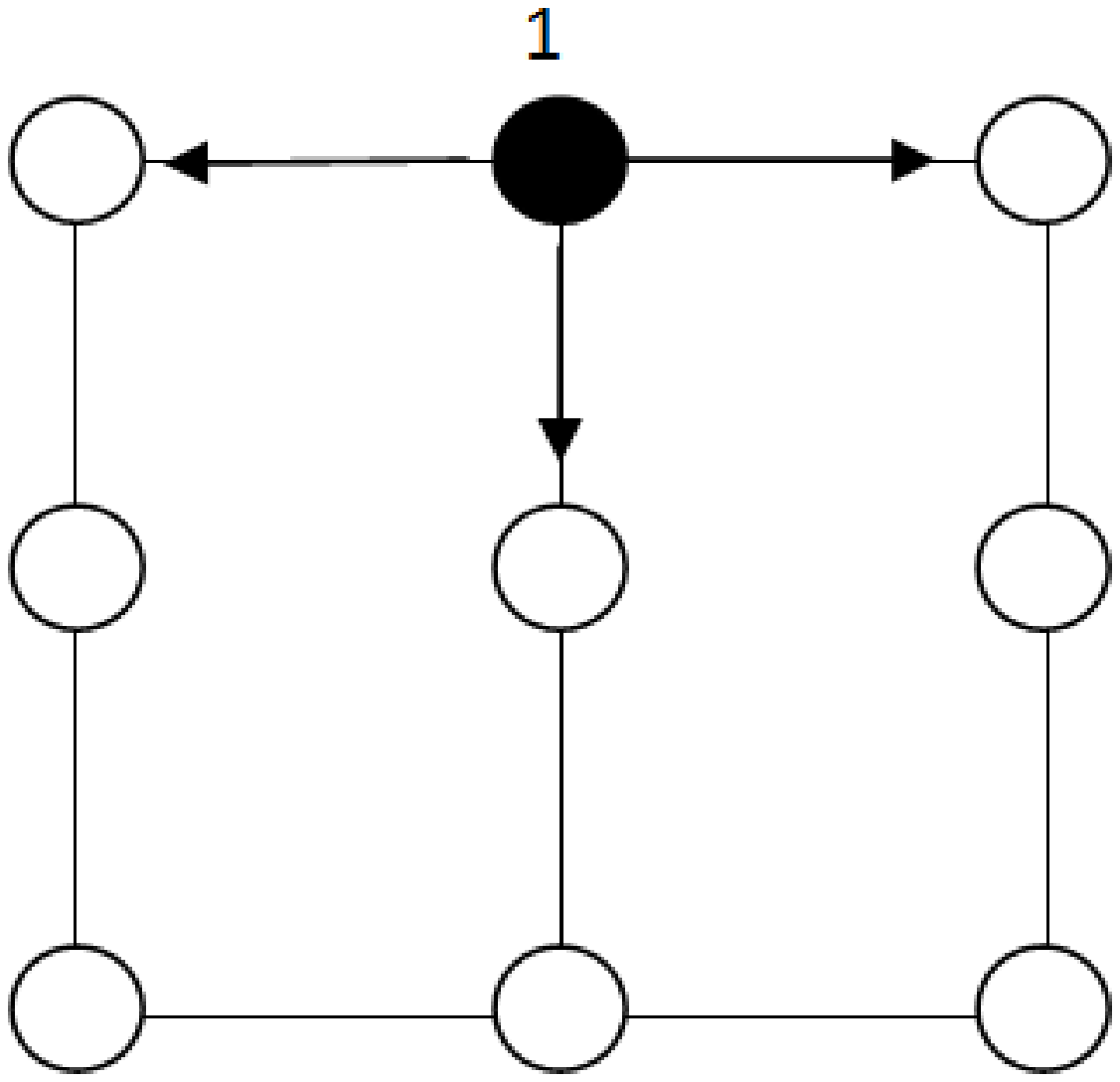}}
\subfigure[First round of BP]{
\label{fig:1-3}
\includegraphics[width=0.3\columnwidth]{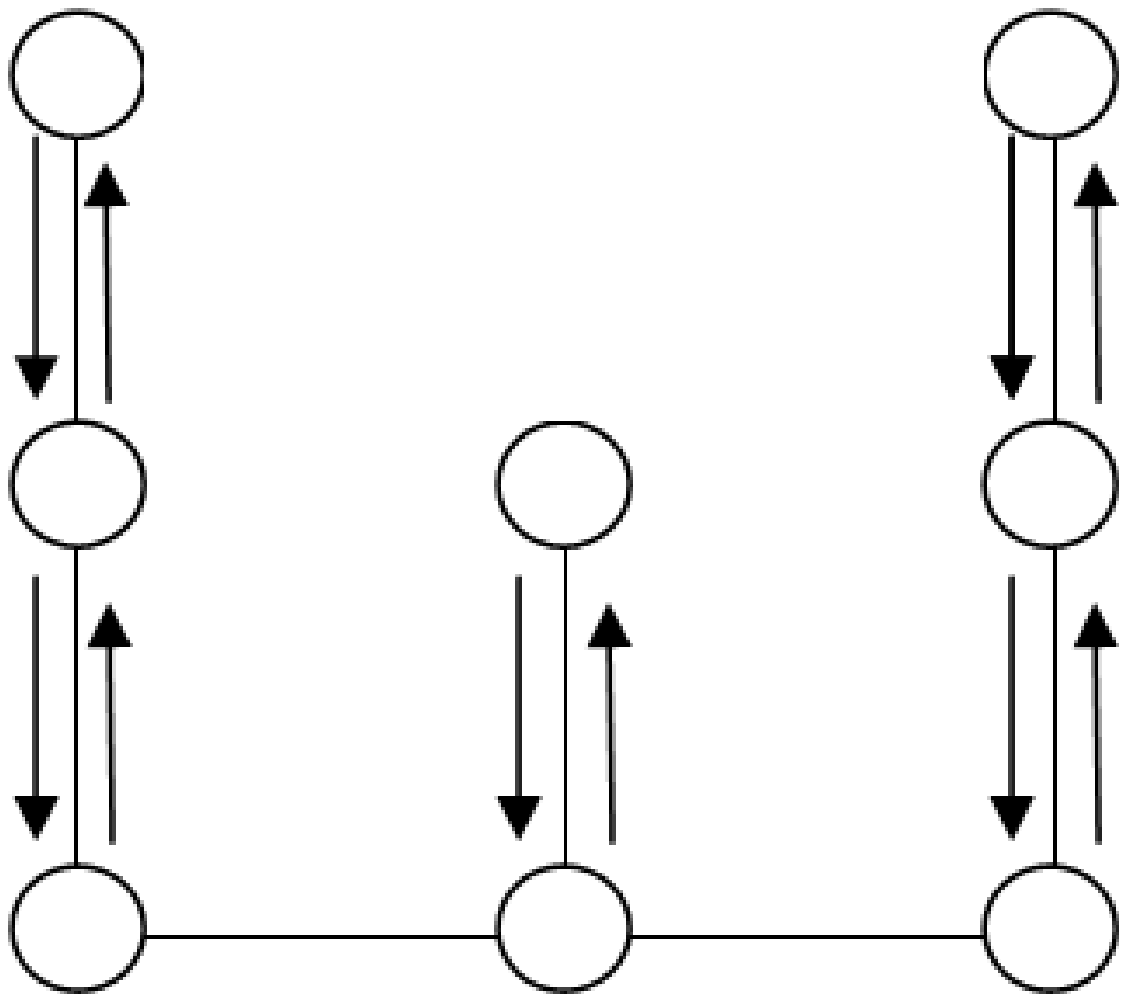}}
\subfigure[Forward messages]{
\label{fig:1-4}
\includegraphics[width=0.3\columnwidth]{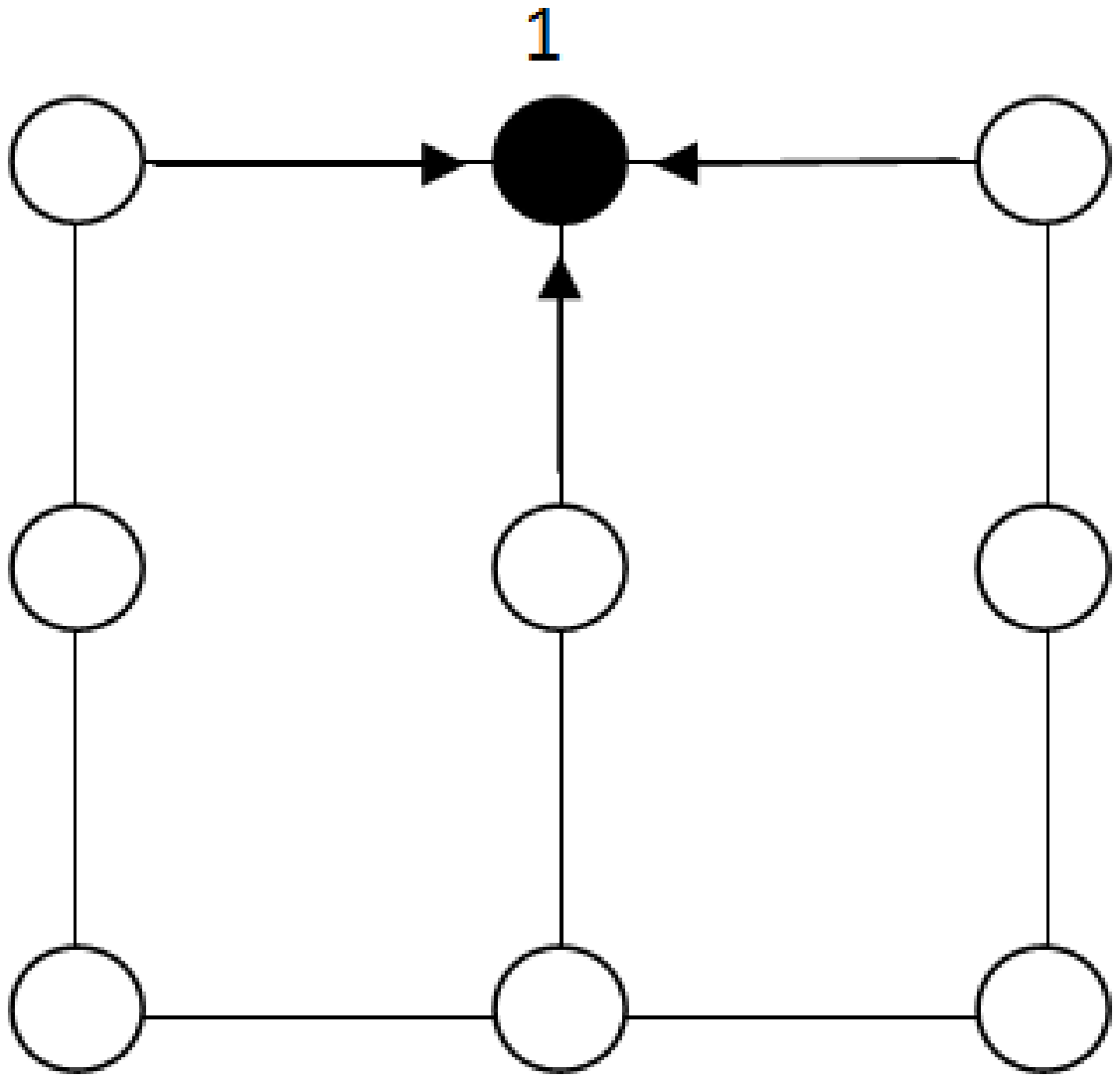}}
\subfigure[Feedback messages]{
\label{fig:1-5}
\includegraphics[width=0.3\columnwidth]{start}}
\subfigure[Second round of BP]{
\label{fig:1-6}
\includegraphics[width=0.3\columnwidth]{bpontree}}
\label{fig:1}
\caption{The FMP algorithm with a single feedback node}
\end{figure}

Consider the loopy graph in Figure \ref{fig:1-1} and a Gaussian graphical model, with information matrix $J$ and potential vector $\bh$, defined on it. Let $J$ and $\bh$ be the information matrix and potential vector of the model respectively. In this graph every cycle passes through node $1$, and thus node $1$ forms an FVS by itself. We use $\calT$ to denote the subgraph excluding node $1$ and its incident edges. Graph $\calT$ is a tree, which does not have any cycles.\footnote{More generally, the cycle-free graph used in FMP can be a collection of disconnected trees, i.e., a forest.} Using node $1$ as the feedback node, FMP consists of the following steps:

{\em{Step 1:}} Initialization

We construct an additional potential vector $\bh^{1}=J_{\calT,1}$ on $\calT$, i.e. $\bh^1$ is the submatrix (column vector) of $J$ with column index $1$ and row indices corresponding to $\calT$. Note that, since in this case $\calF = \{1\}$, this new potential vector is precisely $J_{\calT\calF}$. In the general case $J_{\calT\calF}$ will consist of a set of columns, one for each element of the FVS, where each of those columns is indexed by the nodes in $\calT$. Note that $h^1_i=J_{1i}$, for all $i\in\calN(1)$ and $h^1_i=0$, for all $i\notin \calN(1)\cup\{1\}$. We can view this step as node $1$ sending messages to its neighbors to obtain $\bh^1$. See Figure \ref{fig:1-2} for an illustration.

{\em{Step 2:}} First Round of BP on $J_\calT$ (Figure \ref{fig:1-3})

We now perform BP on $\calT$ twice, both times using the information matrix $J_\calT$, but two different potential vectors. The first of these is simply the original potential vector restricted to $\calT$, i.e., $\bh_\calT$. The second uses $\bh^1$ as constructed in Step 1.\footnote{Note that since both BP passes here -- and, in the general case, the set of $k+1$ BP passes in this step -- use the same information matrix. Hence there are economies in the actual BP message-passing as the variance computations are the same for all.} The result of the former of these BP sweeps yields for each node $i$ in $\calT$ its ``partial variance'' $P^{\calT}_{ii}=(J_{\calT}^{-1})_{ii}$ and its ``partial mean'' $\mu^{\calT}_{i}=(J_{\calT}^{-1}\bh_{\calT})_{i}$ by standard BP message passing on $\calT$. Note that these results are not the true variances and means since this step does not involve the contributions of node $1$. At the same time, BP using $\bh^1$ yields a ``feedback gain'' $g^1_i$, where $g^1_i=(J_{\calT}^{-1}\bh^1)_i$ by standard BP on $\calT$.\footnote{The superscript $1$ of $g^1_i$ means this feedback gain corresponds to the feedback node $1$, notation we need in the general case} Since $\calT$ is a tree-structured graph, BP terminates in linear time.

{\em{Step 3:}} Exact Inference for the Feedback Node

Feedback node $1$ collects the ``feedback gains'' from its neighbors as shown in Figure \ref{fig:1-4}. Node $1$ then calculates its {\em exact} variance and mean as follows:
\begin{eqnarray}
\label{eq:gain0}
P_{11}&=&(J_{11}-\sum_{j\in \calN(1)}{J_{1j}g^{1}_{j}})^{-1},\\
\mu_1&=&P_{11}(h_1-\sum_{j\in \calN(1)}{J_{1j}\mu^{\calT}_{j}}).
\end{eqnarray}

In this step, all the computations involve only the parameters local to node $i$, the ``feedback gains'' from, and the ``partial means'' of node $1$'s neighbors. 

{\em{Step 4:}} Feedback Message Passing (Figure \ref{fig:1-5})

After feedback node $1$ obtains its own variance and mean, it passes the results to all other nodes in order to correct their ``partial variances'' $P^{\calT}_{ii}$ and ``partial means'' $\mu^{\calT}_{i}$ computed in Step 2. 

The neighbors of node $1$ revise their node potentials as follows:
\begin{equation}
\widetilde{h}_j=\left\{
\begin{array}{ll}
h_j-J_{1j}\mu_1,& \forall j\in \calN(1),\\
h_j,& \forall j\notin \calN(1).
\end{array}
\right.
\label{eq:revise}
\end{equation}
From \eqref{eq:revise} we see that only node $1$'s neighbors revise their node potentials. The revised potential vector $\widetilde{\bh}_{\calT}$ and $J_{\calT}$ are then used in the second round of BP. 

{\em{Step 5:}} Second Round of BP on $J_{\calT}$ (Figure \ref{fig:1-6})

We perform BP on $\calT$ with $J_{\calT}$ and $\widetilde{\bh}_{\calT}$ ). The means $\mu_i=(J_{\calT}^{-1}\widetilde{\bh}_{\calT})_i,\hspace{3pt}$ obtained from this round of BP are the {\em exact} means.

The {\em exact} variances can be computed by adding correction terms to the ``partial variances'' as
\begin{equation}
P_{ii}=P^{\calT}_{ii}+P_{11}(g^{1}_{i})^2, \qquad\forall i\in \calT,
\end{equation}
\noindent
where the ``partial variance'' $P^{\calT}_{ii}$  and the ``feedback gain'' $g^{1}_{i}$ are computed in Step $2$. There is only one correction term in this single feedback node case. We will see that when the size of FVS is larger than one, there will be multiple correction terms.

\subsection{Feedback Message Passing for General Graphs}
For a general graph, the removal of a single node may not break all cycles. Hence, the FVS may consist of multiple nodes. In this case, the FMP algorithm for a single feedback node can be generalized by adding extra feedback messages, where each extra message corresponds to one extra feedback node in the FVS. 

Assume an FVS, $\calF$, has been selected, and, as indicated previously, we order the nodes such that $\calF=\{1,\ldots, k\}$. The FMP algorithm with multiple feedback nodes is essentially the same as the FMP algorithm with a single feedback node. When there are $k$ feedback nodes, we compute $k$ sets of feedback gains each corresponding to one feedback node. More precisely, Step 1 in the algorithm now involves performing BP on $\calT$ $k+1$ times, all with the same information matrix, $J_\calT$, but with different potential vectors, namely $\bh_\calT$ and $\bh^p$, $p = 1,…,k$, where these are the successive columns of $J_{\calT\calF}$. To obtain the exact inference results for the feedback nodes, we then need to solve an inference problem on a smaller graph, namely $\calF$, of size $k$, so that Step 3 in the algorithm becomes one of solving a $k$-dimensional linear system. Step 4 then is simply modified from the single-node case to provide a revised potential vector on $\calT$ taking into account corrections from each of the nodes in the FVS. Step 5 then involves a single sweep of BP on $\calT$ using this revised potential vector to compute the exact means on $\calT$, and the feedback gains, together with the variance computation on the FVS, provide corrections to the partial variances for each node in $\calT$. The general FMP algorithm with a given FVS $\calF$ is summarized in Figure \ref{fig:FMP}. 
\begin{figure}
\centering
\noindent
\fbox{\parbox{\columnwidth}{
\noindent
{\bf Input}: information matrix $J$, potential vector $\bh$ and feedback vertex set $\calF$ of size $k$

\noindent
{\bf Output}: mean $\mu_i$ and variance $P_{ii}$ for every node $i$
\begin{enumerate}
\item[1.]
Construct $k$ extra potential vectors: $\forall p\in \calF, \bh^{p}=J_{\calT ,p}$, each corresponding to one feedback node. 
\item[2.]
Perform BP on $\calT$ with $J_{\calT}$, $\bh_{\calT}$ to obtain $P^{\calT}_{ii}=(J_{\calT}^{-1})_{ii}$ and $\mu^{\calT}_{i}=(J_{\calT}^{-1}\bh_{\calT})_i$ for each $i\in\calT$. With the $k$ extra potential vectors, calculate the feedback gains $g^{1}_i=(J_{\calT}^{-1}\bh^1)_i, g^{2}_i=(J_{\calT}^{-1}\bh^2)_i, \ldots, g^{k}_i=(J_{\calT}^{-1}\bh^k)_i$ for $i\in\calT$ by BP.
\item[3.]
Obtain a size-$k$ subgraph with $\widehat{J}_{\calF}$ and $\widehat{\bh}_{\calF}$ given by
\begin{eqnarray*}
(\widehat{J}_\calF)_{pq}&=&J_{pq}-\sum_{j\in \calN(p)\cap \calT}J_{pj}g^{q}_j,\hspace{5pt} \forall p,\hspace{1pt}q\in \calF,\\
(\widehat{\bh}_{\calF})_{p}&=&h_{p}-\sum_{j\in \calN(p)\cap\calT}J_{p j}\mu^{\calT}_j,\hspace{5pt} \forall p\in \calF,
\end{eqnarray*}
and solve the inference problem on the small graph by
$P_{\calF}=\widehat{J}_{\calF}^{-1}$ and $\bmu_{\calF}=\widehat{J}_{\calF}^{-1}\widehat{\bh}_{\calF}$.
\item[4.]
Revise the potential vector on $\calT$ by

\begin{equation}
\tilde{h}_i=h_i-\sum_{j\in\calN(i)\cap \calF}{J_{ij}(\bmu_{\calF})_j}, \hspace{5pt} \forall i\in\calT.\nonumber
\end{equation}
\item[5.]
Another round of BP with the revised potential vector $\widetilde{\bh}_{\calT}$ gives the exact means for nodes on $\calT$. 

Add correction terms to obtain the exact variances for nodes in $\calT$:  

\begin{equation}
P_{ii}=P^{\calT}_{ii}+\sum_{p\in \calF}{\sum_{q\in \calF}{g^{p}_i (P_\calF)_{pq}g^{q}_i}}, \hspace{5pt}\forall i\in{\calT}.\nonumber
\end{equation}

\end{enumerate}
}}
\caption{The FMP algorithm with a given FVS}
\label{fig:FMP}
\end{figure}
\subsection{Correctness and Complexity of FMP}
\label{subsec:ac}
In this subsection, we analyze the correctness and computational complexity of FMP. 
\begin{theorem}
The feedback message passing algorithm described in Figure \ref{fig:FMP} results in the exact means and exact variances for all nodes.
\label{thm:FMP}
\end{theorem}
\begin{IEEEproof}
To make the notation in what follows somewhat less cluttered, let $J_{M}=J_{\calT\calF}$ so that we can write
\begin{equation}
	J=\left[
	\begin{array}{cc}
		J_{\calF}  &  J_M'\\
		J_{M}      &  J_{\calT}
	\end{array}
	\right]\text{  and }
	\bh=\left[
		\begin{array}{c}
			\bh_{\calF}\\
			\bh_{\calT}
		\end{array}
	\right].
\label{eq:partition}
\end{equation}

Similarly, we can write
\begin{equation}
	P=\left[
	\begin{array}{cc}
		P_{\calF}  &  P_M'\\
		P_{M}      &  P_{\calT}
	\end{array}
	\right]\text{  and }
	\bmu=\left[
		\begin{array}{c}
			\bmu_{\calF}\\
			\bmu_{\calT}
		\end{array}
	\right].
\label{eq:patition2}
\end{equation}

By the construction of $\bh^1,\bh^2,\ldots, \bh^k$ in FMP and (\ref{eq:partition}), 
\begin{equation}
	J_M=[\bh^1,\bh^2,\ldots,\bh^k].
\end{equation}

The feedback gains $\bg^1,\bg^2,\ldots, \bg^k$ in FMP are computed by BP with $\bh^1,\bh^2,\ldots,\bh^k$ as potential vectors. Since BP gives the exact means on trees,
\begin{equation}
	[\bg^1,\bg^2,\ldots,\bg^k]=\left[ J_{\calT}^{-1}\bh^1,  J_{\calT}^{-1}\bh^2, \ldots, J_{\calT}^{-1}\bh^k \right]=J_{\calT}^{-1}J_M.
\label{eq:gain}
\end{equation}
In FMP, $\bmu^{\calT}$ is computed by BP with potential vector $\bh_{\calT}$, so
\begin{equation}
	\bmu^{\calT}=J_{\calT}^{-1}\bh_{\calT}.
\label{eq:meanmean}
\end{equation}
\noindent The diagonal of $J_{\calT}^{-1}$ is also calculated exactly in the first round of BP in FMP as $P^{\calT}_{ii}=(J_{\calT}^{-1})_{ii}$. 

Since $P=J^{-1}$, by matrix computations, we have
\begin{equation}
P_{\calT}=J_{\calT}^{-1}+(J_{\calT}^{-1}J_M)P_{\calF}(J_{\calT}^{-1}J_M)'.
\label{eq:correction}
\end{equation}
Substituting (\ref{eq:gain}) into (\ref{eq:correction}), we have
\begin{equation}
	P_{ii}=P^{\calT}_{ii}+\sum_{p\in \calF}\sum_{q\in \calF}{g^p_i\left( P_{\calF} \right)_{pq} g^q_i}, \quad \forall i\in \calT,
\end{equation}
\noindent where $P^{\calT}_{ii}$ is the ``partial variance'' of node $i$ and $g^p_i$ the ``feedback gain'' in FMP. Here $P_{\calF}$ is the exact covariance matrix of the feedback nodes in $\calF$. This is the same equation as in Step $5$ of FMP. We need to show that $P_{\calF}$ is indeed calculated exactly in FMP. 

By Schur's complement,
\begin{equation}
\widehat{J}_{\calF} \deff P_{\calF}^{-1}=J_{\calF}-J_M'J_{\calT}^{-1}J_M \hspace{5pt}\text{ and }\hspace{5pt}
\widehat{\bh}_{\calF} \deff P_{\calF}^{-1}\bmu_{\calF}=\bh_{\calF}-J_M'J_{\calT}^{-1}\bh_{\calT}.
\end{equation}

By \eqref{eq:gain} and \eqref{eq:meanmean},
\begin{equation}
	\widehat{J}_{\calF}=J_{\calF}-J_M'[\bg^1,\bg^2,\ldots,\bg^k]
	\hspace{5pt}\text{ and }\hspace{5pt}\widehat{\bh}_{\calF}=\bh_{\calF}-J_M'\bmu^{\calT},
\end{equation}
\noindent which is exactly the same formula as in Step 3 of FMP.
Therefore, we obtain the exact covariance matrix and exact means for nodes in $\calF$ by solving $P_{\calF}=(\widehat{J}_{\calF})^{-1}$ and $\bmu_\calF=P_{\calF}\widehat{\bh}_\calF$.

Since $\bmu=J^{-1}\bh$, from (\ref{eq:partition}) and (\ref{eq:patition2}) we can get
\begin{equation}
\bmu_{\calT}=J_{\calT}^{-1}(\bh_{\calT}-J_M\bmu_{\calF}).
\end{equation}

We define $\widetilde{\bh}_{\calT}=\bh_{\calT}-J_M\bmu_{\calF}$, i.e.,
\begin{equation}
	(\widetilde{\bh}_{\calT})_i=h_i-\sum_{j\in\calN(i)\cap \calF}{J_{ij}(\mu_\calF)_j},
\end{equation}
\noindent where $\bmu_{\calF}$ is the exact mean of nodes in $\calF$. This step is equivalent to performing BP with parameters $J_{\calT}$ and the revised potential vector $\widetilde{\bh}_{\calT}$ as in Step 4 of FMP. This completes the proof.
\end{IEEEproof}
We now analyze the computational complexity of FMP with $k$ denoting the size of the FVS and $n$ the total number of nodes in the graph. In Step 1 and Step 2, BP is performed on $\calT$ with $k+2$ messages (one for $J$, one with $\bh_\calT$, and one for each $\bh^p$). The total complexity is $\calO(k(n-k))$. In step 3, $\calO(k^2(n-k))$ computations are needed to obtain $\widehat{J}_{\calF}$ and $\widehat{h}_{\calF}$ and $\calO(k^3)$ operations to solve the inference problem on a graph of size $k$. In Step 4 and Step 5, it takes $\calO(k(n-k))$ computations to give the exact means and $\calO(k^2(n-k))$ computations to add correction terms. Therefore, the total complexity is $\calO(k^2n)$. Therefore, the computational complexity of FMP is $\calO(k^2n)$. This is a significant reduction from $\calO(n^3)$ of direct matrix inversion when $k$ is small. 


\section{Approximate Feedback Message Passing}
\label{chap:approximateFMP} 
As we have seen from Theorem \ref{thm:FMP}, FMP always gives correct inference results. However, FMP is intractable if the size of the FVS is very large. This motivates our development of {\em approximate FMP}, which uses a pseudo-FVS instead of an FVS.
\subsection{Approximate FMP with a Pseudo-FVS}
\label{sec:Approx1a}
There are at least two steps in FMP which are computationally intensive when $k$, the size of the FVS, is large: solving a size-$k$ inference problem in Step 3 and adding $k^2$ correction terms to each non-feedback node in Step 5. One natural approximation is to use a set of feedback nodes of smaller size. We define a {\em pseudo-FVS} as a subset of an FVS that does not break all the cycles. A useful pseudo-FVS has a small size, but breaks the most ``crucial'' cycles in terms of the resulting inference errors. We will discuss how to select a good pseudo-FVS in Section \ref{sec:Approx1c}. In this subsection, we assume that a pseudo-FVS is given.

Consider a Gaussian graphical model Markov on a graph $\calG=(\calV,\calE)$. We use $\pF$ to denote the given pseudo-FVS, and use $\pT$ to denote the pseudo-tree (i.e., a graph with cycles) obtained by eliminating nodes in $\pF$ from $\calG$. With a slight abuse of terminology, we still refer to the nodes in $\pF$ as the feedback nodes. A natural extension is to replace BP by LBP in Step 2 and Step 5 of FMP.\footnote{Of course, one can insert other algorithms for Steps 2 and 5 -- e.g., iterative algorithms such as embedded trees \cite{chandrasekaran2008estimation} which can yield exact answers. However, here we focus on the use of LBP for simplicity.}  

The total complexity of approximate FMP depends on the size of the graph, the cardinality of the pseudo-FVS, and the number of iterations of LBP within the pseudo-tree. Let $k$ be the size of the pseudo-FVS, $n$ be the number of nodes, $m$ be the number of edges in the graph, and $D$ be the maximum number of iterations in Step 2 and Step 5. By a similar analysis as for FMP, the total computational complexity for approximate FMP is $\calO(k^2n+kmD)$. Assuming that we are dealing with relatively sparse graphs, so that $m = \calO(n)$, reductions in complexity as compared to a use of a full FVS rely on both $k$ and $D$ being of moderate size. Of course the choices of those quantities must also take into account the tradeoff with the accuracy of the computations.
\subsection{Convergence and Accuracy}
\label{sec:Approx1b}

In this subsection, we provide theoretical results on convergence and accuracy of approximate FMP. We first provide a result assuming convergence that makes several crucial points, namely on the exactness of means throughout the entire graph, the exactness of variances on the pseudo-FVS, and on the interpretation of the variances on the remainder of the graph as augmenting the LBP computation with a rich set of additional walks, roughly speaking those that go through the pseudo-FVS:
\begin{theorem}
Consider a Gaussian graphical model with parameters $J$ and $\bh$. If approximate FMP converges with a pseudo-FVS $\pF$, it gives the correct means for all nodes and the correct variances on the pseudo-FVS. The variance of node $i$ in $\pT$ calculated by this algorithm equals the sum of all the backtracking walks of node $i$ within $\pT$ plus all the self-return walks of node $i$ that visit $\pF$, so that the only walks missed in the computation of the variance at node $i$ are the non-backtracking walks within $\pT$.  
\label{thm:Fcorrect}
\end{theorem}

\begin{IEEEproof}
We have
\begin{equation}
	J=\left[
	\begin{array}{cc}
		J_{\pF}  &  J_M'\\
		J_{M}      &  J_{\pT}
	\end{array}
	\right]\text{  and }
	\bh=\left[
		\begin{array}{c}
			\bh_{\pF}\\
			\bh_{\pT}
		\end{array}
	\right].
\end{equation} 

By Result \ref{prop:LBPconverge}) in Section \ref{sec:walksum}, when LBP converges, it gives the correct means. Hence, after convergence, for $i=1,2,\ldots,k$, we have
\begin{equation}
\nonumber \bg^i=J_{\pT}^{-1}J_{\pT,i},\hspace{5pt} \text{ and } \hspace{5pt} \bmu^{\pT}=J_{\pT}^{-1}\bh_{\pT},
\label{eq:F2}
\end{equation}
\noindent where $\bg^i$ is the feedback gain corresponding to feedback node $i$ and $\bmu^{\pT}$ is the partial mean in approximate FMP. These quantities are exact after convergence.  

Since  $\bg^i$ and $\bmu^{\pT}$ are computed exactly, following the same steps as in the proof of Theorem \ref{thm:FMP}, we can obtain the exact means and variances for nodes in $\pF$.

From the proof of Theorem \ref{thm:FMP}, we also have 
\begin{equation}
\bmu_{\pT}=J_{\pT}^{-1}(\bh_{\pT}-J_M\bmu_{\pF}).
\label{eq:Tmeancorrect}
\end{equation}
 
We have shown that $\bmu_{\pF}$ is computed exactly in Step 3 in approximate FMP, so $\bh_{\pT}-J_M\bmu_{\pF}$ is computed exactly. Since LBP on $\pT$ gives the exact means for any potential vector, the means of all nodes in $\pT$ are exact. 

As in the proof of Theorem \ref{thm:FMP}, we have that the exact covariance matrix on $\pT$ is given by 
\begin{equation}
P_{\pT}=J_{\pT}^{-1}+(J_{\pT}^{-1}J_M)P_{\pF}(J_{\pT}^{-1}J_M)'.
\label{eq:Tapprox}
\end{equation}
As noted previously, the exact variance of node $i\in\pT$ equals the sum of all the self-return walks of node $i$. We partition these walks into two classes: self-return walks of node $i$ within $\pT$, and self-returns walks that visit at least one node in $\pF$. The diagonal of $J_{\pT}^{-1}$ captures exactly the first class of walks. Hence, the second term in the right-hand side of (\ref{eq:Tapprox}) corresponds to the sum of the second class of walks. Let us compare each of these terms to what is computed by the approximate FVS algorithm. By Result \ref{prop:backtracking}) in Section \ref{sec:walksum}, LBP on $\pT$ gives the sums of all the backtracking walks after convergence. So the first term in (\ref{eq:Tapprox}) is approximated by backtracking walks. However, note that the terms $J_{\pT}^{-1}J_M$ and $P_{\pF}$ are obtained exactly.\footnote{Note that the columns of the former are just the feedback gains computed by LBP for each of the additional potential vectors on $\pT$ corresponding to columns of $J_{\pT\pF}$, which we have already seen are computed exactly, as we have for the covariance on the pseudo-FVS.} Hence, the approximate FMP algorithm computes the second term exactly and thus provides precisely the second set of walks. As a result, the only walks missing from the exact computation of variances in $\pT$ are non-backtracking walks within $\calT$. This completes the proof.
\end{IEEEproof}
We now state several conditions under which we can guarantee convergence.
\begin{proposition}
Consider a Gaussian graphical model with graph $\calG=(\calV,\calE)$ and model parameters $J$ and $\bh$. If the model is walk-summable, approximate FMP converges for any pseudo-FVS $\pF\subset\calV$.
\label{prop:FMPconverge}
\end{proposition}
\begin{IEEEproof}
Let $R=I-J$ and $(\bar{R})_{ij}=|R_{ij}|$. 
In approximate FMP, LBP is performed on the pseudo-tree induced by $\pT=\calV\backslash\pF$. The information matrix on the pseudo-tree is $J_{\pT}$, which is a submatrix of $J$. By Corollary 8.1.20 in \cite{horn1990matrix} 
, for any $\pT$
\begin{equation}
\rho(\bar{R}_{\pT})\leq\rho(\bar{R})<1.
\label{eq:spectralradius}
\end{equation}

By Result \ref{prop:LBPconverge}) in Section \ref{sec:walksum}, LBP on $\pT$ is guaranteed to converge. All other computations in approximate FMP terminate in a finite number of steps. Hence, approximate FMP converges for any pseudo-FVS $\pF\subset\calV$.
\end{IEEEproof}      

For the remainder of the paper we will refer to the quantities as in (\ref{eq:spectralradius}) as the spectral radii of the corresponding graphs (in this case $\pT$ and the original graph $\calG$). Walk-summability on the entire graphical model is actually far stronger than is needed for approximate FMP to converge. As the proof of Proposition \ref{prop:FMPconverge} suggests, all we really need is for the graphical model on the graph excluding the pseudo-FVS to be walk-summable. As we will discuss in Section \ref{sec:Approx1c}, this objective provides one of the drivers for a very simple algorithm for choosing a pseudo-FVS in order to enhance the walk-summability of the remaining graph and as well as accuracy of the resulting LBP variance computations.

{\bf Remarks:} The following two results follow directly from Proposition \ref{prop:FMPconverge}.
\begin{enumerate}
\item
Consider a walk-summable Gaussian graphical model. Let $\pF_j$ be a pseudo-FVS consisting of $j$ nodes and $\emptyset\neq\pF_1\subseteq\pF_2\subseteq\cdots\subseteq \pF_k\subseteq \calF$, where $\calF$ is an FVS, then $W^{\mathrm{LBP}}_i\subseteq W^{\pF_1}_i\subseteq W^{\pF_2}_i\subseteq\ldots\subseteq W^{\pF_k}_i\subseteq W^{\calF}_i$ for any node $i$ in the graph. Here $W^{\mathrm{LBP}}_i$ is the set of walks captured by LBP for calculating the variance of node $i$; $W^{\pF_j}_i$ is the set of walks captured by approximate FMP with pseudo-FVS $\pF_j$; and $W^{\calF}_i$ is the set of walks captured by FMP with FVS $\calF$.
\label{cor:FMPwalks}
\item
Consider an attractive Gaussian graphical model (i.e., one in which all elements of $R$ are non-negative). Let $\pF_1\subseteq \pF_2 \subseteq \cdots \subseteq \pF_k \subseteq \calF$ denote the pseudo-FVS (FVS), and $P^{\mathrm{LBP}}_{ii}$, $P^{\pF_1}_{ii}$, $\ldots$, $P^{\pF_k}_{ii}$, $P^{\calF}_{ii}$ denote the corresponding variances calculated for node $i$ by LBP, approximate FMP and FMP respectively. $P_{ii}$ represents the exact variance of node $i$. We have 
$P^{\mathrm{LBP}}_{ii}\leq P^{{\pF}_1}_{ii}\leq P^{{\pF}_2}_{ii} \leq \cdots \leq P^{{\pF}_k}_{ii}\leq P^{\calF}_{ii}=P_{ii}$ for any node $i$ in $\calV$.
\label{thm:FMPattractive}
\end{enumerate}

The above results show that with approximate FMP, we can effectively trade off complexity and accuracy by selecting pseudo-FVS of different sizes. 
\subsection{Error Bounds for Variance Computation}
\label{sec:errorbounds}
We define the measure of the error of an inference algorithm for Gaussian graphical models as the average absolute error of variances for all nodes:
\begin{equation}
\epsilon=\frac{1}{n}\sum_{i\in\calV}{|\widehat{P}_{ii}-P_{ii}|},
\label{eq:errordef}
\end{equation}
\noindent where $n$ is the number of nodes, $\widehat{P}_{ii}$ is the computed variance of node $i$ by the algorithm and $P_{ii}$ is the exact variance of node $i$.
\begin{proposition}
Consider a walk-summable Gaussian graphical model with $n$ nodes. Assume the information matrix $J$ is normalized to have unit diagonal. Let $\epsilon_{\mathrm{FMP}}$ denote the error of approximate FMP and $\widehat{P}^{\mathrm{FMP}}_{ii}$ denote the estimated variance of node $i$. Then  
\begin{equation}
\nonumber
\epsilon_ {\mathrm{FMP}}=\frac{1}{n}\sum_{i\in\calV}{|\widehat{P}^{\mathrm{FMP}}_{ii}-P_{ii}|}\leq  \frac{n-k}{n}\frac{\tilde{\rho}^{\tilde{g}}}{1-\tilde{\rho}},
\end{equation}
\noindent where $k$ is the number of feedback nodes, $\tilde{\rho}$ is the spectral radius corresponding to the subgraph $\widetilde{\calT}$, and $\tilde{g}$ denotes the girth of $\widetilde{\calT}$, i.e., the length of the shortest cycle in $\widetilde{\calT}$. In particular, when $k=0$, i.e., LBP is used on the entire graph, we have  
\begin{equation}
\nonumber
\epsilon_{\mathrm{LBP}}=\frac{1}{n}\sum_{i\in\calV}{|\widehat{P}^{\mathrm{LBP}}_{ii}-P_{ii}|}\leq \frac{\rho^g}{1-\rho},
\end{equation}
\noindent where the notation is similarly defined. 
\label{thm:FMPerror}
\end{proposition}
\newcommand{\abswalksumlong}[3]{\bar{\phi}(#1\stackrel{#3}{\longrightarrow}#2)}

Some of the following proof techniques are motivated by the proof of the error bound on determinant estimation with the so-called orbit-product representation in \cite{johnson2009orbit}.
\begin{IEEEproof}
By Theorem \ref{thm:Fcorrect},
\begin{equation}
\epsilon_{\mathrm{LBP}}=\frac{1}{n}\sum_{i\in\calV}|\walksumlong{i}{i}{\text{NB}}|,
\end{equation}
\noindent  where $\walksumlong{i}{i}{\text{NB}}$ denotes the sum of all non-backtracking self-return walks of node $i$.

We have 
\begin{equation}
\epsilon_{\mathrm{LBP}}=\frac{1}{n}\sum_{i\in\calV}|\walksumlong{i}{i}{\text{NB}}|\leq \frac{1}{n}\sum_{i\in\calV}\abswalksumlong{i}{i}{\text{NB}}, 
\end{equation}
\noindent where $\bar{\phi}(\cdot)$ denotes the sum of absolute weight of walks, or walk-sums defined on $\bar{R}$. 

Non-backtracking self-return walks must contain at least one cycle. So the minimum length of a non-backtracking walk is $g$, which is the minimum length of cycles. Thus
\begin{align}
\epsilon_{\mathrm{LBP}}&\leq \frac{1}{n}\sum_{i\in\calV}\abswalksumlong{i}{i}{\text{NB}}
				 \le \frac{1}{n}\sum_{i\in\calV}\sum_{m=g}^{\infty}(\bar{R}^m)_{ii}\\
				 &=\frac{1}{n}\mathrm{Tr}(\sum_{m=g}^{\infty}(\bar{R}^m))
				 =\frac{1}{n}\sum_{m=g}^{\infty}\mathrm{Tr}(\bar{R}^m).
\end{align}
Let $\lambda_i(\cdot)$ denotes the $i$th largest eigenvalue of a matrix. Since $\lambda_i(\bar{R}^m)=\lambda_i(\bar{R})^m$ and $\lambda_i(\bar{R})\leq \rho$, we have
\begin{equation}
\mathrm{Tr}(\bar{R}^m)=\sum_{i=1}^{n}\lambda_i(\bar{R})^m\leq n\rho^m.
\end{equation}
Therefore,
\begin{equation}
\epsilon_{\mathrm{LBP}}\leq\frac{1}{n}\sum_{m=g}^{\infty}{n\rho^m}
             = \frac{\rho^g}{1-\rho}.
\end{equation}

When approximate FMP is used with a size-$k$ pseudo-FVS, the variances of nodes in the pseudo-FVS are computed exactly, while the variance errors for other nodes are the same as performing LBP on the subgraph excluding the pseudo-FVS. Therefore,
\begin{align}
\epsilon_{\mathrm{FMP}}&=\frac{1}{n}\sum_{i\in\calV} {|\hat{P}_{ii}-P_{ii}|}=\frac{1}{n}\sum_{i\in \pT}{|\hat{P}_{ii}-P_{ii}|}\\
&= \frac{1}{n}(n-k)\epsilon_{\mathrm LBP}\leq  \frac{n-k}{n}\frac{\tilde{\rho}^{\tilde{g}}}{1-\tilde{\rho}}.
\end{align}
\end{IEEEproof}

An immediate conclusion of Proposition \ref{thm:FMPerror} is that if a graph is cycle-free (i.e., $g = \infty$), the error $\epsilon_{\mathrm{LBP}}$ is zero. 

We can also analyze the performance of FMP on a Gaussian graphical model that is Markov on a Erd{\H{o}}s-R\'{e}nyi random graph $\mathfrak{G}(n, c/n)$. Each edge in such a random graph with $n$ nodes appears with probability $c/n$, independent of every other edge in the graph \cite{bollobás2001random}. 
\begin{proposition}
Consider a sequence of graphs $\{\calG_n\}_{n=1}^{\infty}$ drawn from Erdos-Renyi model $\mathfrak{G}(n,c/n)$ with fixed $c$. Suppose we have a sequence of Gaussian graphical models parameterized by $\{(J_n,\bh_n)\}_{n=1}^{\infty}$ that are Markov on $\{\mathfrak{G}_n\}_{n=1}^{\infty}$ and are strictly walk-summable (i.e., the spectral radii $\rho(\bar{R}_n)$ are uniformly upper bounded away from unity). Then asymptotically almost surely there exists a sequence of pseudo-FVS $\{\pF_n\}_{n=1}^{\infty}$ with $\pF_n$ of size $\calO(\log n)$, with which the error of approximate FMP as in \eqref{eq:errordef} approaches zero. 
\end{proposition}
\begin{IEEEproof}
We can obtain a graph with girth greater than $l$ by removing one node at every cycle of length up to $l$. The number of cycles of length up to $l$ in $\mathfrak{G}(n, c/n)$ is $\calO(c^l)$ asymptotically almost surely (Corollary 4.9 in \cite{bollobás2001random}). So we can obtain a graph of girth $\log \log n$ by removing $\calO(\log n)$ nodes. By Proposition \ref{thm:FMPerror}, the error approaches zero when $n$ approaches infinity.
\end{IEEEproof}
 

\subsection{Finding a good Pseudo-FVS of Bounded Size}
\label{sec:Approx1c}
One goal of choosing a good pseudo-FVS is to ensure that LBP converges on the remaining subgraph; the other goal is to obtain smaller inference errors. In this subsection we discuss a local selection criterion motivated by these two goals and show that the two goals are consistent.

Let $\bar{R}$ denote the absolute edge weight matrix. Since $\rho(\bar{R})<1$ is a sufficient condition for LBP to converge on graph $\calG$, obtaining convergence reduces to that of removing the minimum number of nodes such that $\rho(\bar{R}_{\pT})<1$ for the remaining graph $\pT$. However, searching and checking this condition over all possible sets of pseudo-FVS's up to a desired cardinality is a prohibitively expensive, and instead we seek a local method (i.e., using only quantities associated with individual nodes) for choosing nodes for our pseudo-FVS, one at a time, to enhance convergence. The principal motivation for our approach is the following bound \cite{horn1990matrix} on the spectral radius of a nonnegative matrix:
\begin{equation}
\min_{i}{\sum_{j}{\bar{R}_{ij}}}\leq \rho(\bar{R}) \leq \max_{i}{\sum_{j}{\bar{R}_{ij}}}.
\label{eq:bound}
\end{equation}
We further simplify this problem by a greedy heuristic: one feedback node is chosen at each iteration. This provides a basis for a simple greedy method for choosing nodes for our pseudo-FVS. In particular, at each stage, we examine the graph excluding the nodes already included in the pseudo-FVS and select the node with the largest sum of edge weights, i.e., $\underset{i}{\operatorname{argmax}}{\quad \sum_j{\bar{R}_{ij}}}$. We then remove the node from the graph and put it into $\pF$. We continue the same procedure on the remaining graph until the maximum allowed size $k$ of $\tilde{\calF}$ is reached or the remaining graph does not have any cycles. 

\begin{figure}[t]
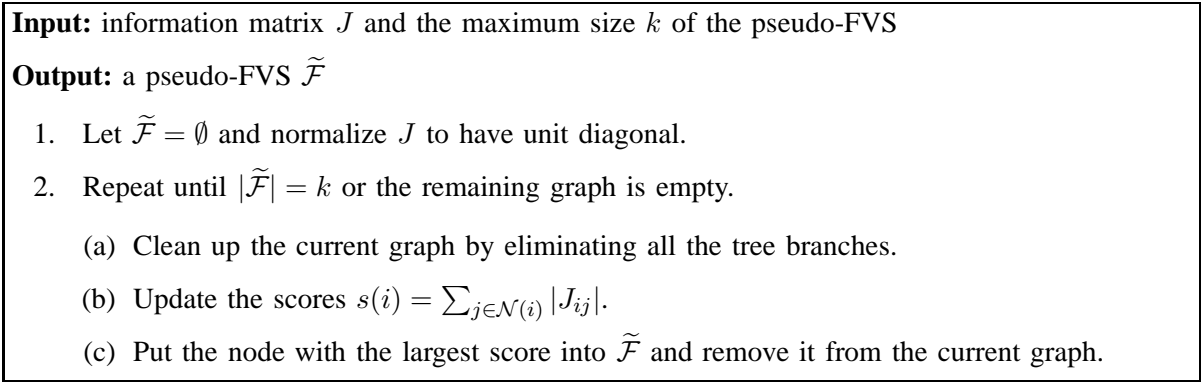

\centering
\fbox{\parbox{0.95\columnwidth}{
\noindent
{\bf Input:} information matrix $J$ and the maximum size $k$ of the pseudo-FVS\\
{\bf Output:} a pseudo-FVS $\pF$
\begin{enumerate}
\item[1. ] Let $\pF=\emptyset$ and normalize $J$ to have unit diagonal.
\item[2. ] Repeat until $|\pF|=k$ or the remaining graph is empty.

\begin{itemize}
\item[(a)]
Clean up the current graph by eliminating all the tree branches. 
\item[(b)]
Update the scores $s(i)=\sum_{j\in\calN(i)}{|J_{ij}|}$. 
\item[(c)]
Put the node with the largest score into $\pF$ and remove it from the current graph.
\end{itemize}
\end{enumerate}
}}
\caption{The pseudo-FVS selection criterion}
\label{fig:FVS1}
\end{figure}

The selection algorithm is summarized in Figure \ref{fig:FVS1}. Note that while the motivation just given for this method is to enhance convergence of LBP on $\pT$, we are also enhancing the accuracy of the resulting algorithm, as Proposition \ref{thm:FMPerror} suggests, since the spectral radius $\rho(\bar{R})$ is reduced with the removal of nodes. In addition, as shown in Theorem \ref{thm:Fcorrect}, the only approximation our algorithm makes is in the computation of variances for nodes in $\pT$, and those errors correspond to non-backtracking self-return walks confined to $\pT$ (i.e., we do capture non-backtracking self-return walks that exit $\pT$ and visit nodes in the pseudo-FVS). Thus, as we proceed with our selection of nodes for our pseudo-FVS, it makes sense to nodes with the largest edge-weights to nodes that are left in $\pT$, which is precisely what this approach accomplishes.

The complexity of the selection algorithms is $\calO(km)$, where $m$ is the number of edges and $k$ is the size of the pseudo FVS. As a result, constructing a pseudo-FVS in this manner is computationally simple and negligible compared to the inference algorithm that then exploits it.

Finding a suitable pseudo-FVS is important. We will see in Section \ref{chap:experiments} that there is a huge performance difference between a good selection and a bad selection of $\tilde{\calF}$. In addition, experimental results show that with a good choice of pseudo-FVS (using the algorithms just described), we not only can get excellent convergence and accuracy results but can do this with pseudo-FVS of cardinality $k$ and number of iterations $D$ that scale well with the graph size $n$. Empirically, we find that we only need $\calO(\log n)$ feedback nodes as well as very few iterations to obtain excellent performance, and thus the complexity is $\calO(n\log^2(n))$.


\section{Numerical Results}
\label{chap:experiments}
In this section, we apply approximate FMP to graphical models that are Markov on two-dimensional grids and present results detailing the convergence and correctness of our proposed algorithm. Two-dimensional grids are sparse since each node is connected to a maximum of four neighbors. There have been many studies of inference problems on grids \cite{george1973nested}. However, inference cannot, in general, be solved exactly in linear time due to the existence of many cycles of various lengths. It is known that the size of the FVS for a grid grows linearly with the number of nodes on the grid \cite{madelaine2008improved}. Hence, we use approximate FMP with a pseudo-FVS of bounded size to ensure that inference is tractable.  

In our simulations, we consider $l\times l$ grids with different values of $l$. The size of the graph is thus $n=l^2$. We randomly generate an information matrix $J$ that has the sparsity pattern corresponding to a grid. Its nonzero off-diagonal entries are drawn from an {\em i.i.d.} uniform distribution with support in $[-1,1]$. We ensure $J$ is positive definite by adding $\lambda I$ for sufficiently large $\lambda$. We also generate a potential vector $\bh$, whose entries are drawn {\em i.i.d.} from a uniform distribution with support in $[-1,1]$. Without loss of generality, we then normalize the information matrix to have unit diagonal.
\subsection{Convergence of Approximate FMP}

\begin{figure}
\centering
\includegraphics[width=0.88\columnwidth]{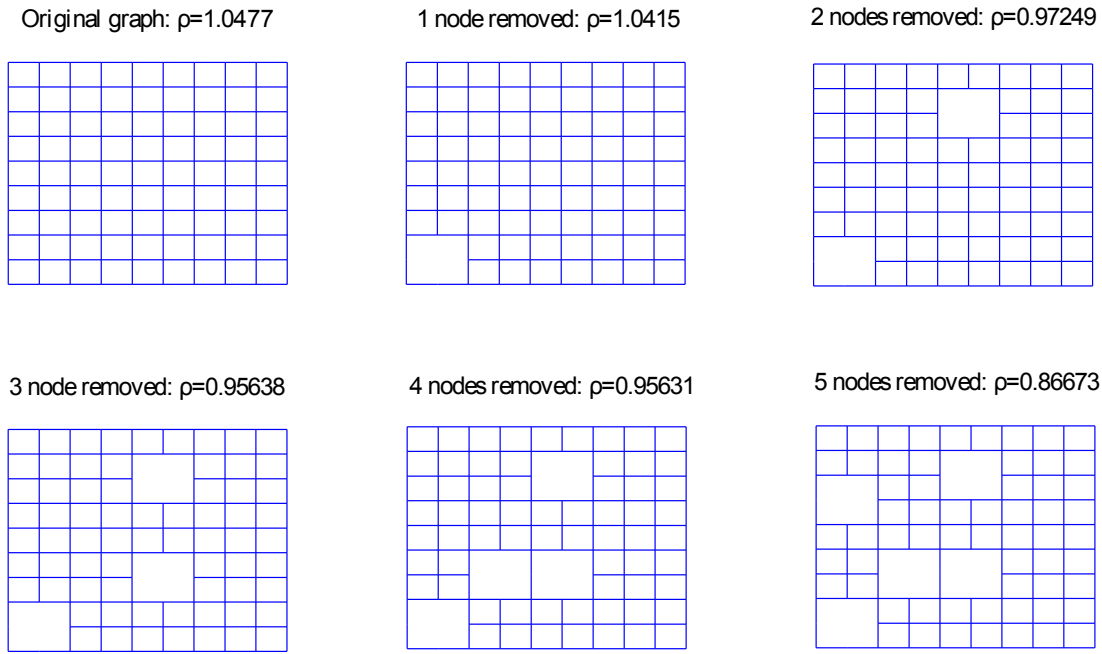}

\caption{Size of the pseudo-FVS and the spectral radius of the corresponding remaining graph}
\label{fig:remove1}
\end{figure}

In Figure \ref{fig:remove1}, we illustrate our pseudo-FVS selection procedure to remove one node at a time for a graphical model constructed as just-described on a $10\times 10$ grid. The remaining graphs, after removing $0$, $1$, $2$, $3$, $4$, and $5$ nodes, and their corresponding spectral radii $\rho(\bar{R})$ are shown in the figures. LBP does not converge on the entire graph and the corresponding spectral radius is $\rho(\bar{R})=1.0477$. When one feedback node is chosen, the spectral radius corresponding to the remaining graph is reduced to $1.0415$. After removing one more node from the graph, the spectral radius is further reduced to $0.97249$, which ensures convergence. In all experiments on $10\times 10$ grids, we observe that by choosing only a few nodes (at most three empirically) for our pseudo-FVS, we can obtain convergence even if LBP on the original graph diverges.  

\begin{figure}
\centering
\subfigure[$10\times 10$ grid]{
\label{fig:spectral-10}
\includegraphics[width=0.44\columnwidth]{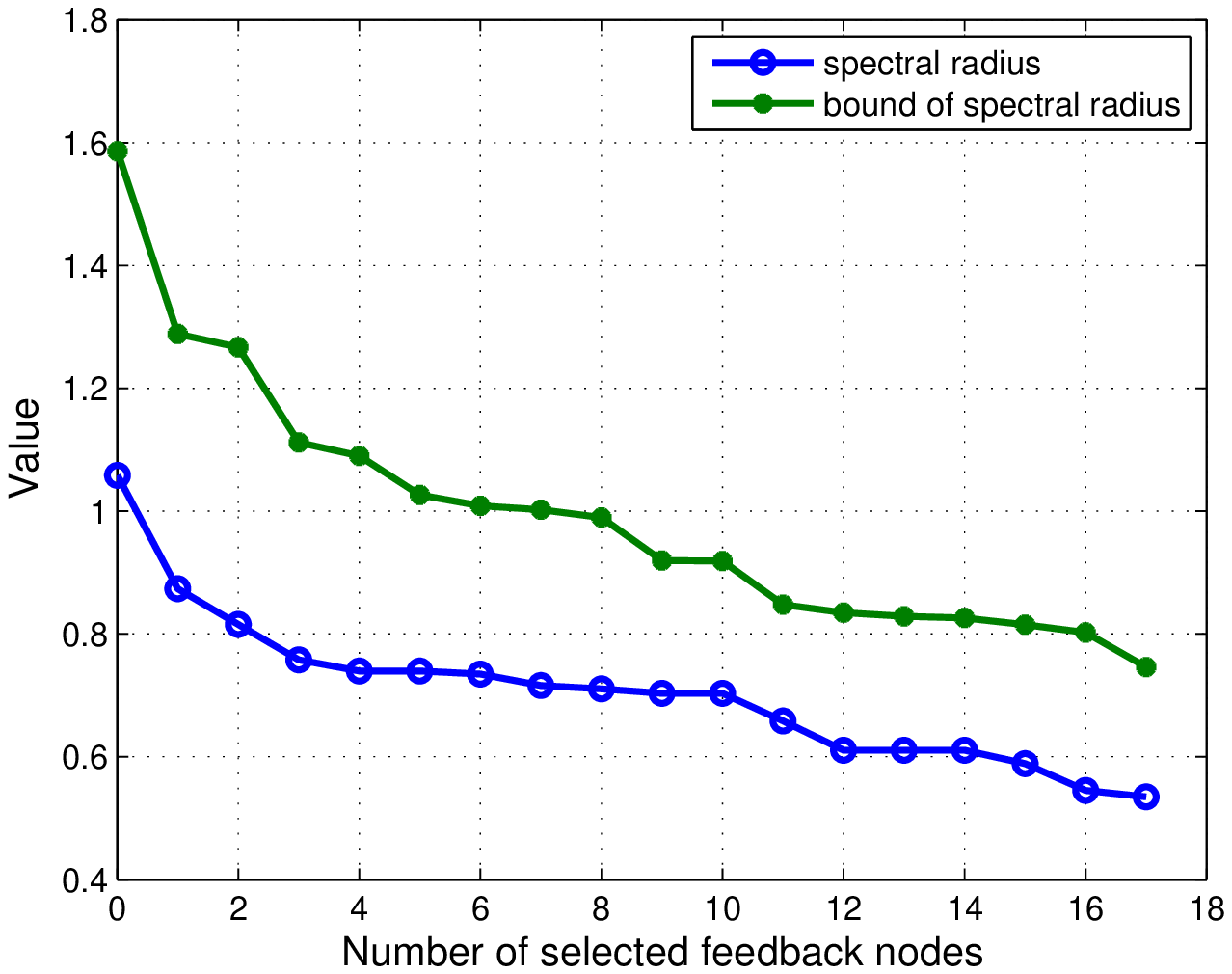}
}
\subfigure[$20\times 20$ grid]{
\label{fig:spectral-20}
\includegraphics[width=0.44\columnwidth]{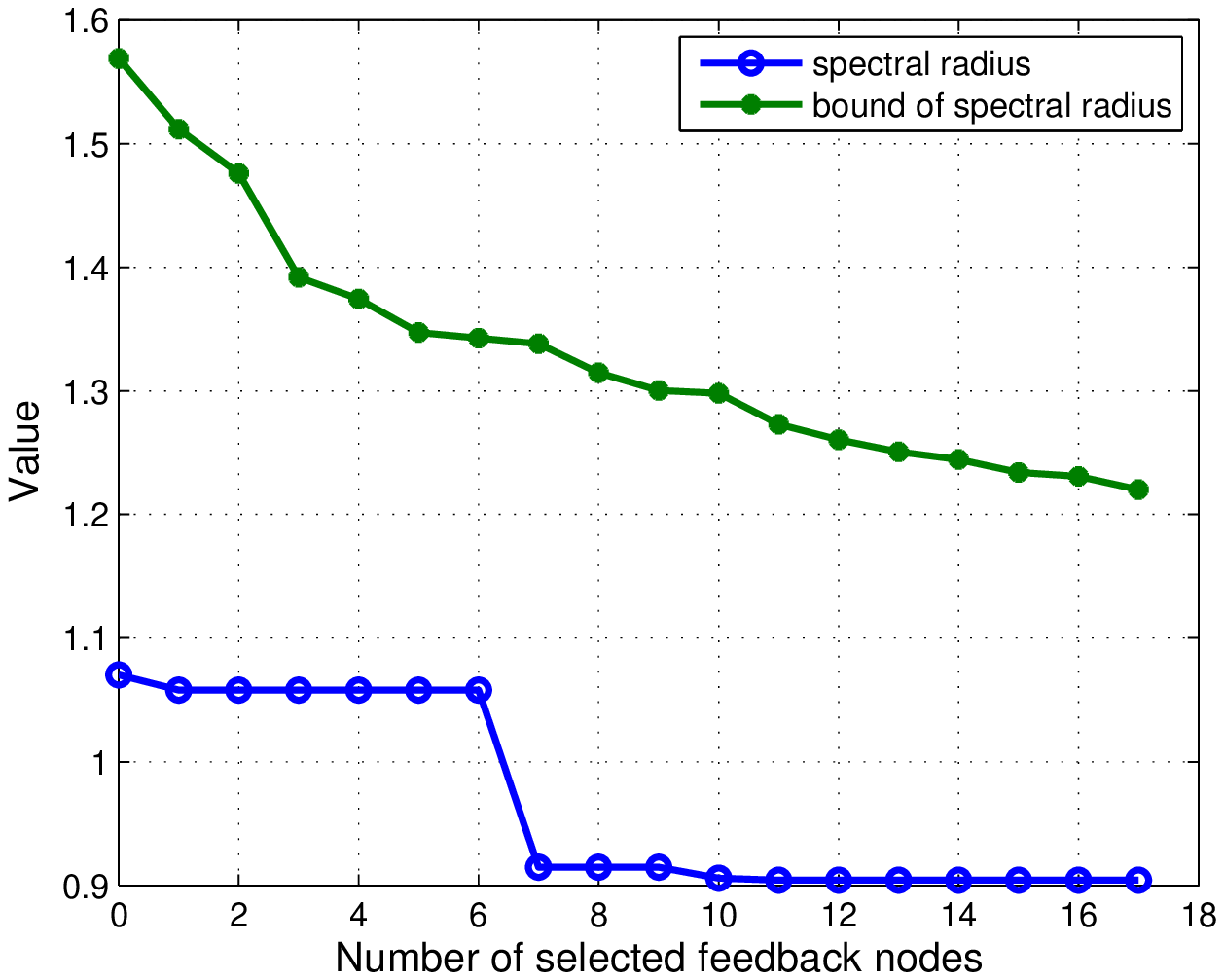}
}
\subfigure[$40\times 40$ grid]{
\label{fig:spectral-40}
\includegraphics[width=0.44\columnwidth]{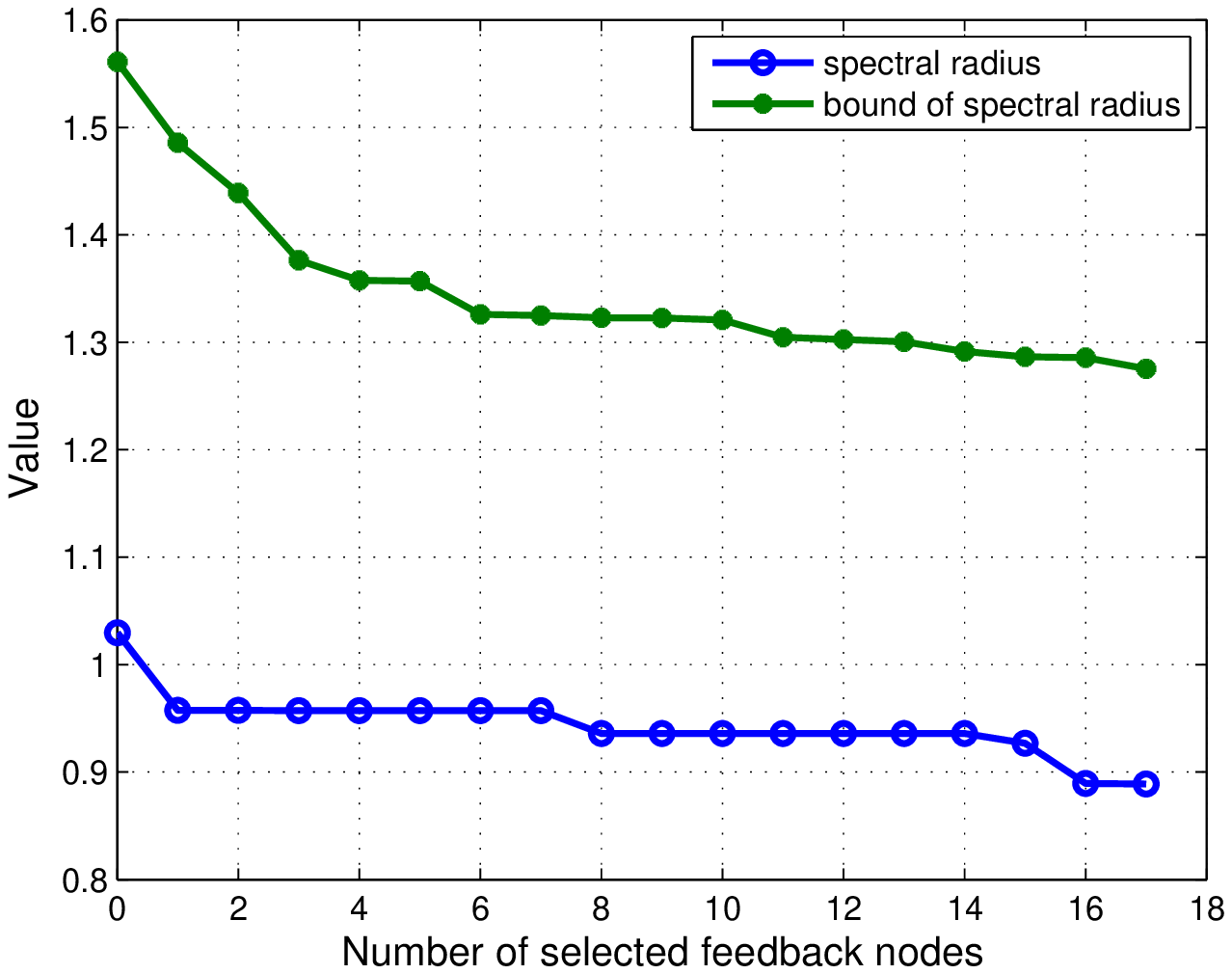}
}
\subfigure[$80\times 80$ grid]{
\label{fig:spectral-80}
\includegraphics[width=0.44\columnwidth]{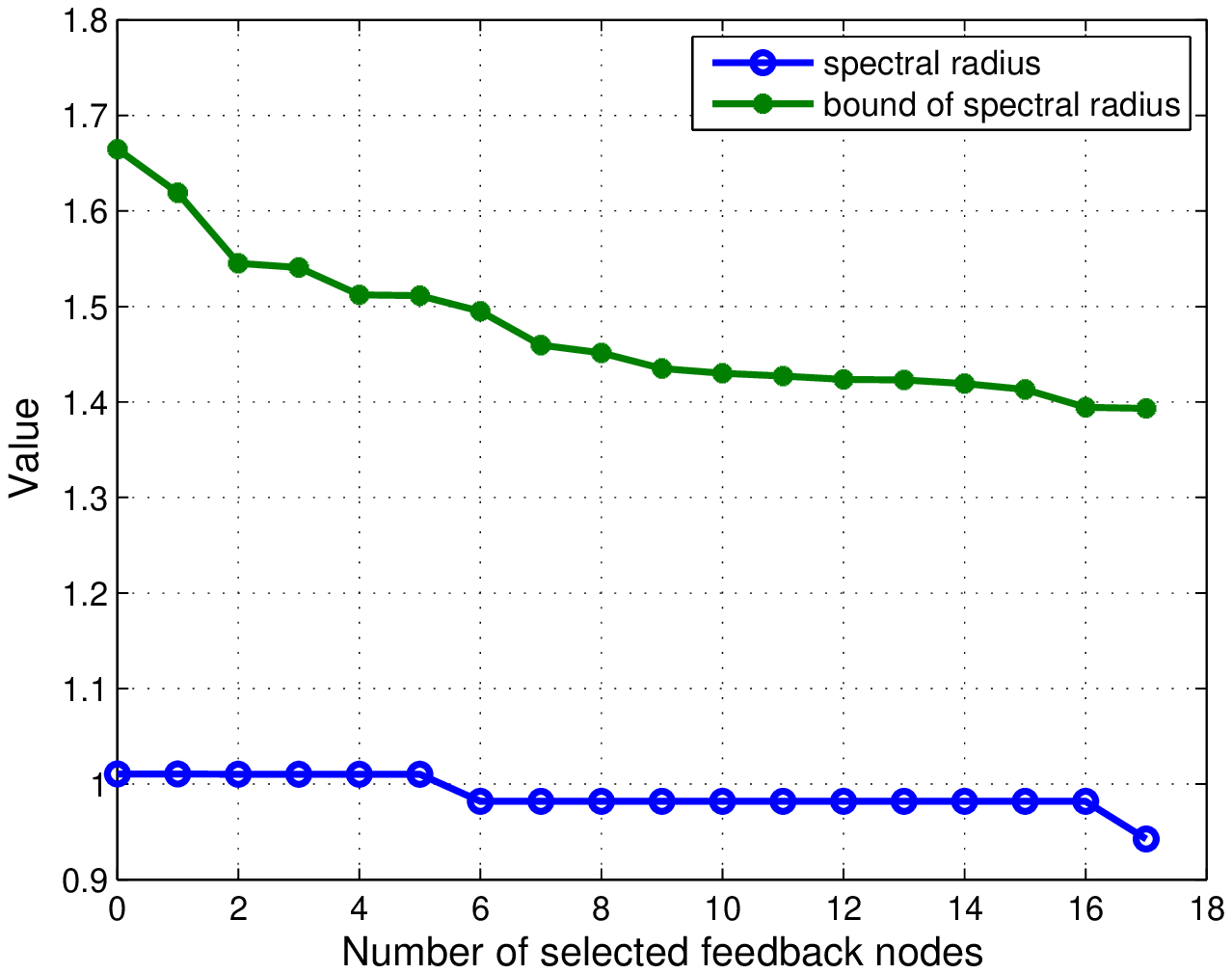}
}
\caption{Number of selected feedback nodes v.s. the spectral radius and its bound}
\label{fig:spectral}
\end{figure}

%
In Figure \ref{fig:spectral} we show that the spectral radius and its upper bound given in (\ref{eq:bound}) decrease when more nodes are included in the pseudo-FVS. Convergence of approximate FMP is immediately guaranteed when the spectral radius is less than one.    

\subsection{Accuracy of Approximate FMP}

In this subsection, we show numerical results of the inference errors defined in (\ref{eq:errordef}). On each grid, LBP and the approximate FMP algorithms with two different sets of feedback nodes are performed. One set has $k=\lceil \log n \rceil$ feedback nodes while the other has $k=\sqrt{n}$ feedback nodes. The horizontal axis shows the number of message passing iterations. The vertical axis shows the errors for both variances and means on a logarithmic scale.\footnote{The error of means is defined in the manner as variances -- the average of the absolute errors of means for all nodes.} 

\begin{figure}
\begin{minipage}{\columnwidth}
\centering
\subfigure[Evolution of variance errors with iterations]{
\label{fig:10-1}
\psfrag{^Iterations^}[c][c][1]{\small Iterations}
\psfrag{^Logoferror^}[c][c][1]{\small Log of error}
\psfrag{^Errorofvariancesfor10by10grid^}[c][c][1]{\small Error of variances for $10\times 10$ grid}
\includegraphics[width=0.47\columnwidth]{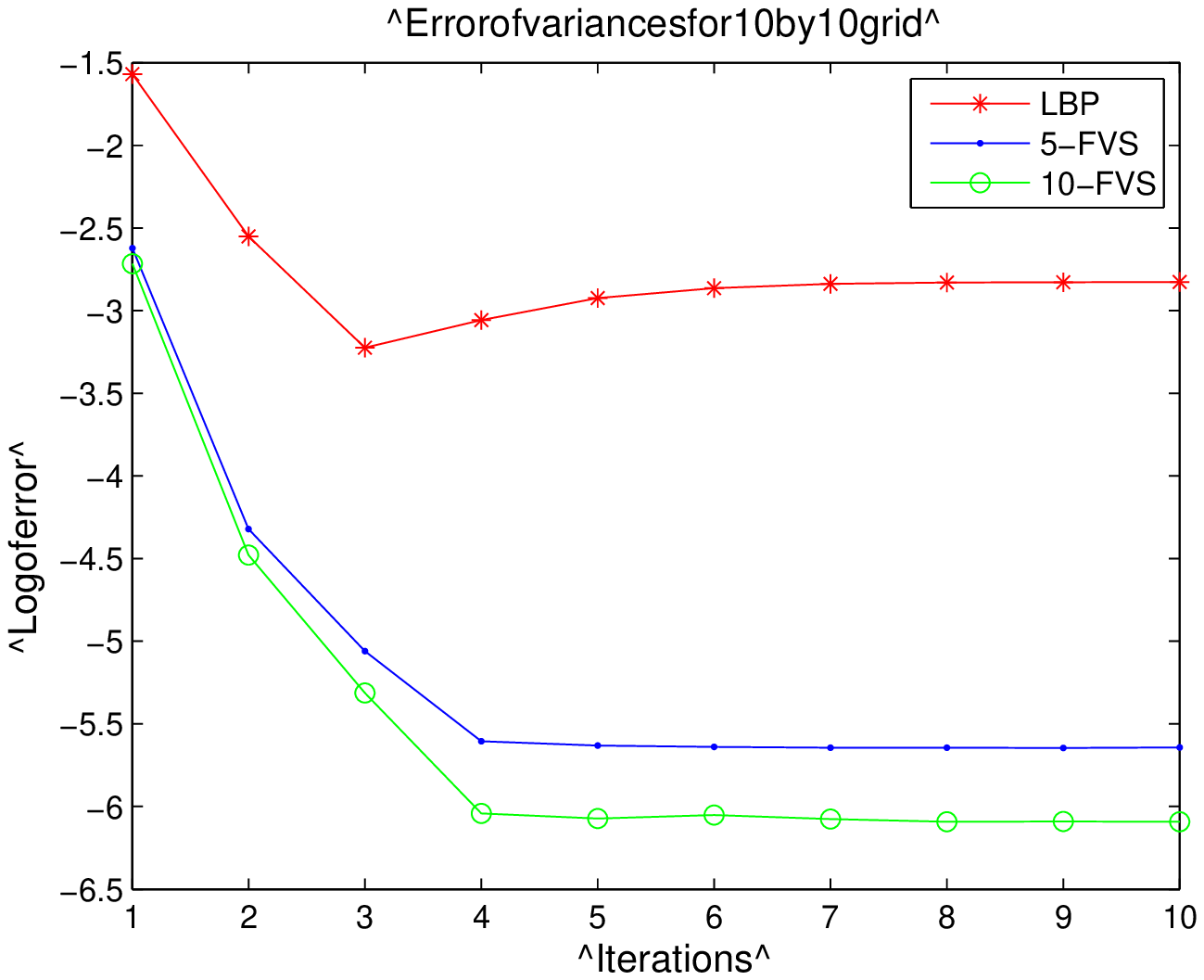}}
\subfigure[Evolution of mean errors with iterations]{
\label{fig:10-2}
\psfrag{^Iterations^}[c][c][1]{\small Iterations}
\psfrag{^Logoferror^}[c][c][1]{\small Log of error}
\psfrag{^Errorofmeansfor10by10grid^}[c][c][1]{\small Error of means for $10\times 10$ grid}
\includegraphics[width=0.47\columnwidth]{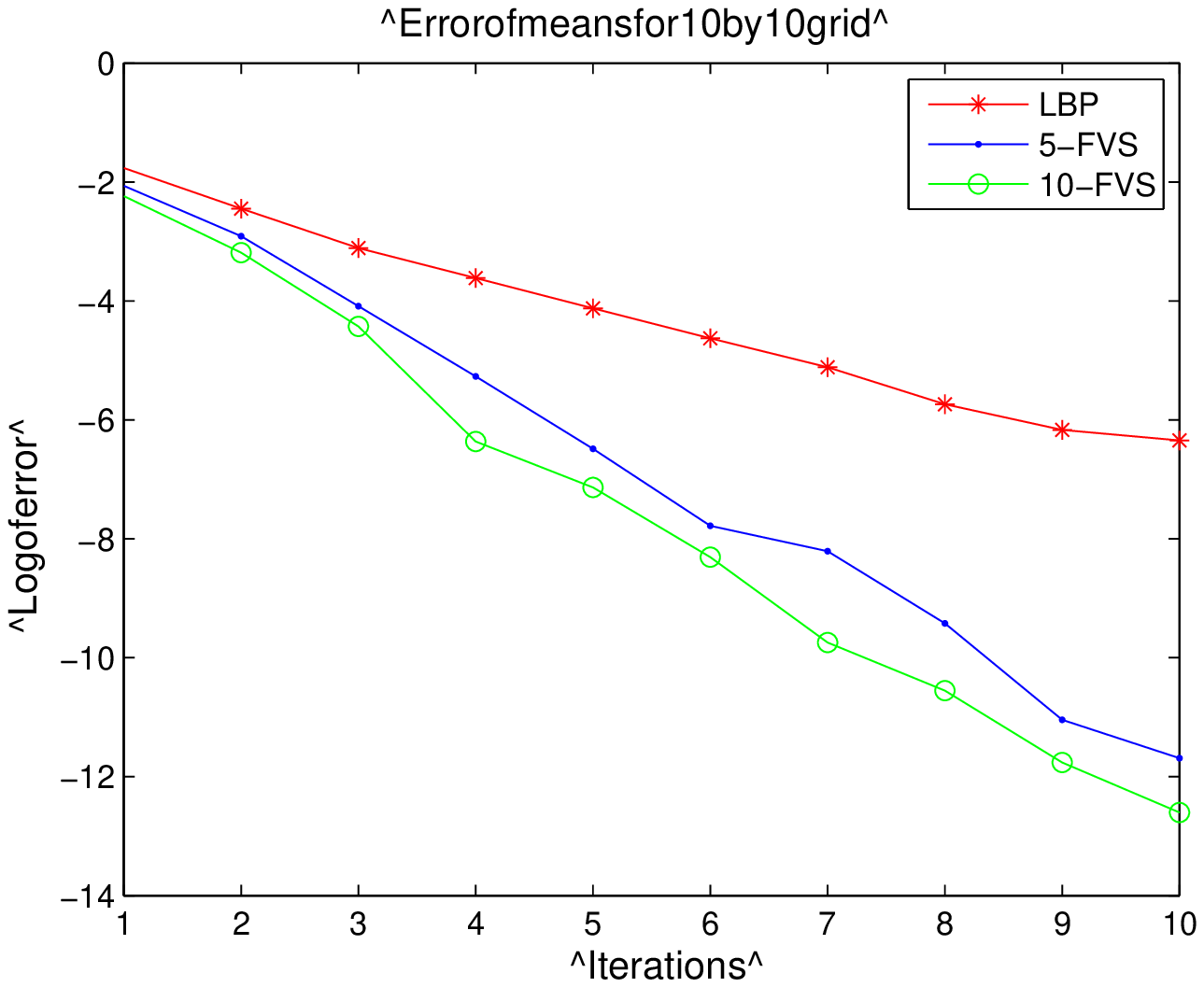}}
\caption{Inference errors of a $10\times 10$ grid}
\label{fig:good10a}
\end{minipage}

\begin{minipage}{\columnwidth}  
\centering
\vspace{15pt}
\subfigure[Evolution of variance errors with iterations]{
\label{fig:10-3}
\psfrag{^Iterations^}[c][c][1]{\small Iterations}
\psfrag{^Logoferror^}[c][c][1]{\small Log of error}
\psfrag{^Errorofvariancesfor10by10grid^}[c][c][1]{\small Error of variances for $10\times 10$ grid}
\includegraphics[width=0.47\columnwidth]{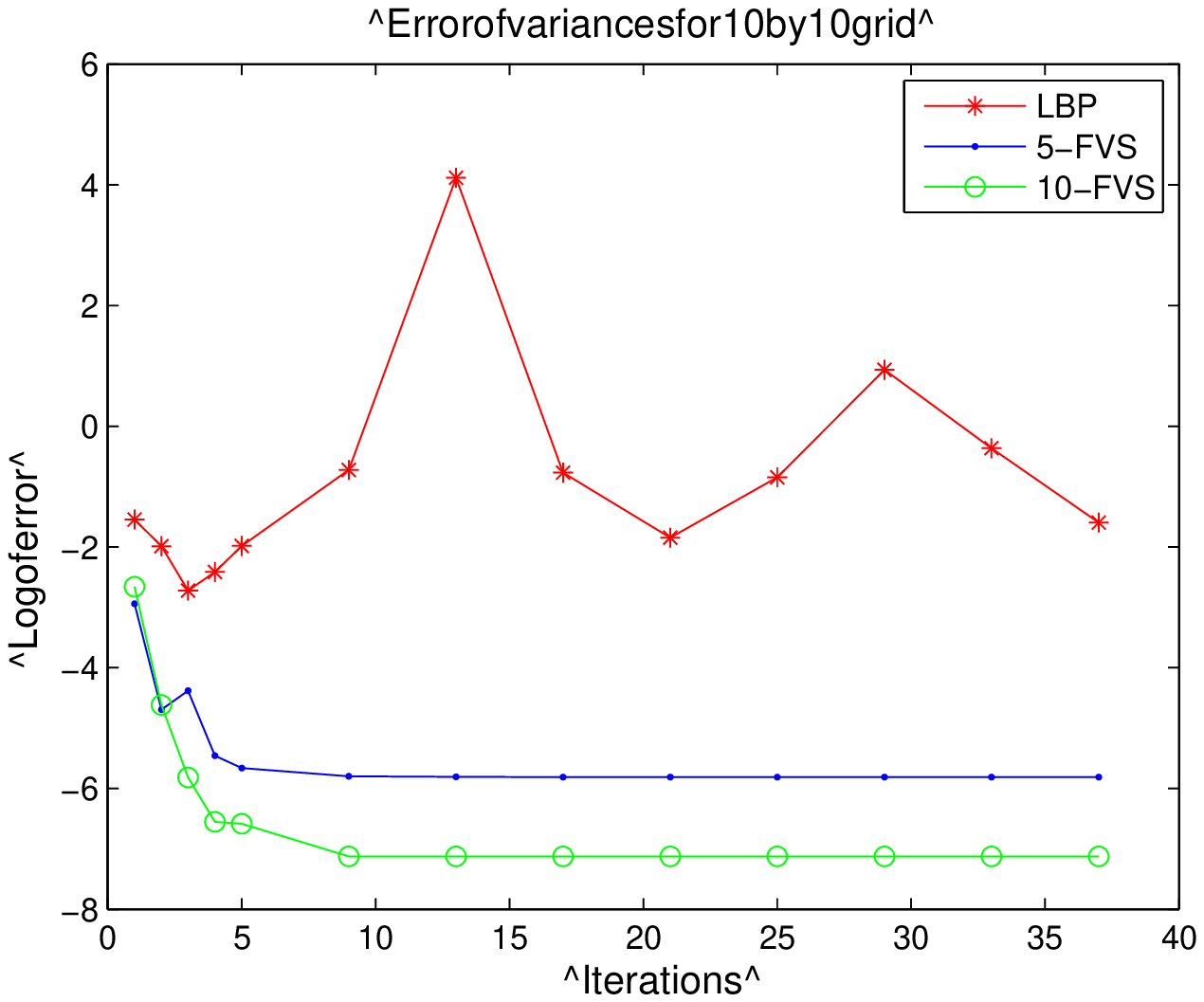}}
\subfigure[Evolution of mean errors with iterations]{
\label{fig:10-4}
\psfrag{^Iterations^}[c][c][1]{\small Iterations}
\psfrag{^Logoferror^}[c][c][1]{\small Log of error}
\psfrag{^Errorofmeansfor10by10grid^}[c][c][1]{\small Error of means for $10\times 10$ grid}
\includegraphics[width=0.47\columnwidth]{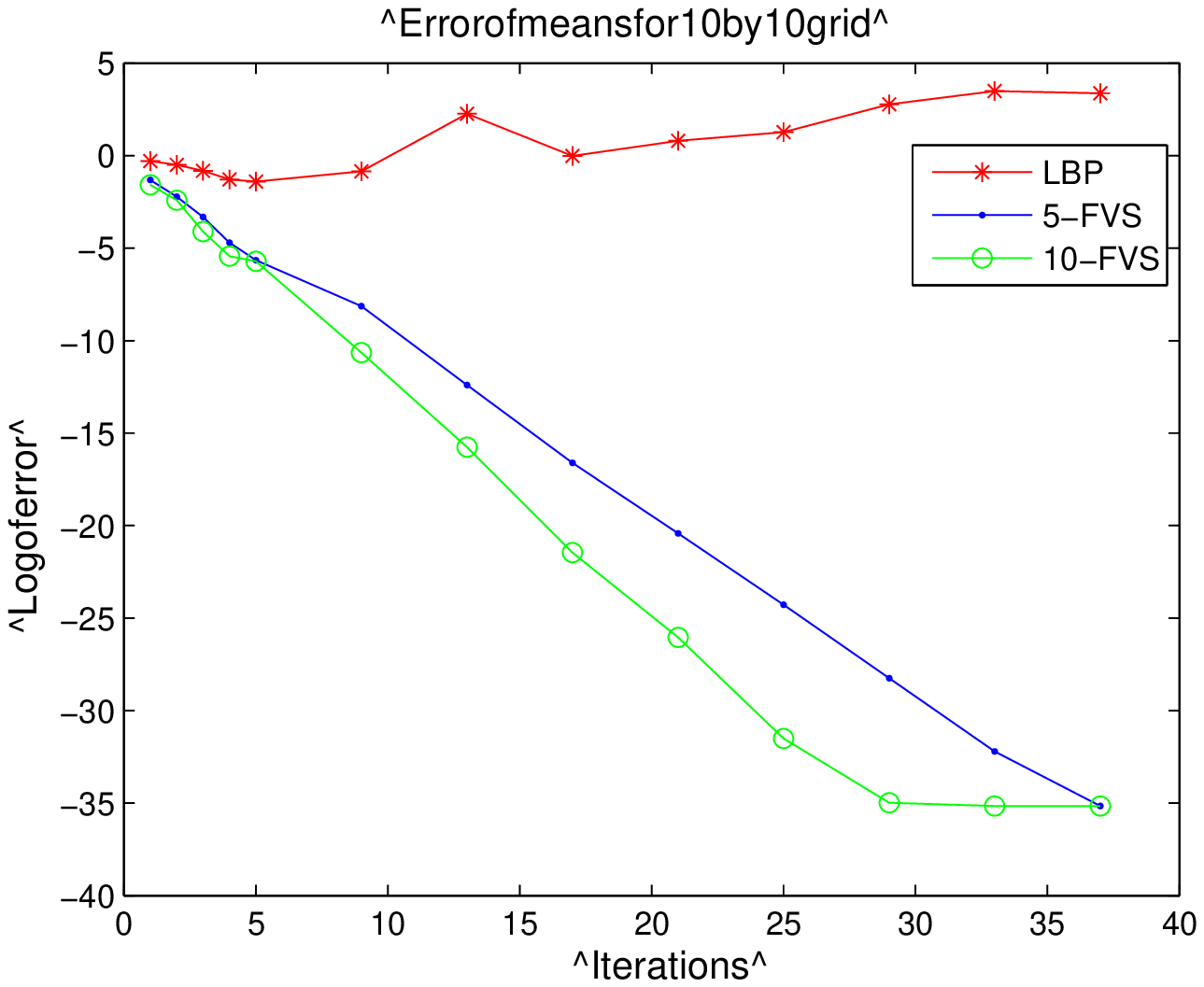}}
\caption{Inference errors of a $10\times 10$ grid}
\label{fig:good10b}
\end{minipage} 
\end{figure}

\begin{figure}
\begin{minipage}{\columnwidth}
\centering
\vspace{-15pt}
\subfigure[Evolution of variance errors with iterations]{
\label{fig:20-1}
\psfrag{^Iterations^}[c][c][1]{\small Iterations}
\psfrag{^Logoferror^}[c][c][1]{\small Log of error}
\psfrag{^Errorofvariancesfor20by20grid^}[c][c][1]{\small Error of variances for $20\times 20$ grid}
\includegraphics[width=0.47\columnwidth]{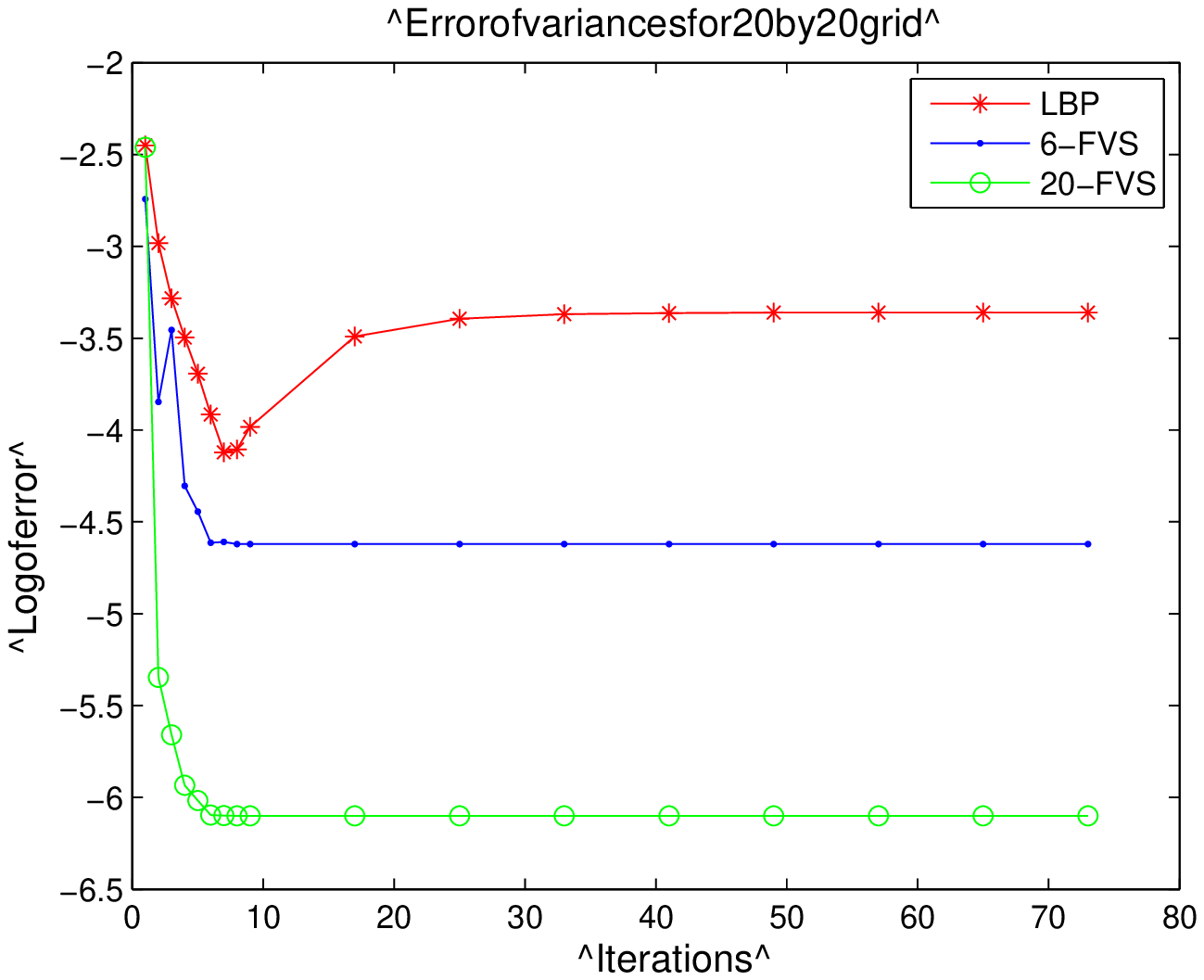}}
\subfigure[Evolution of mean errors with iterations]{
\label{fig:20-2}
\psfrag{^Iterations^}[c][c][1]{\small Iterations}
\psfrag{^Logoferror^}[c][c][1]{\small Log of error}
\psfrag{^Errorofmeansfor20by20grid^}[c][c][1]{\small Error of means for $20\times 20$ grid}
\includegraphics[width=0.47\columnwidth]{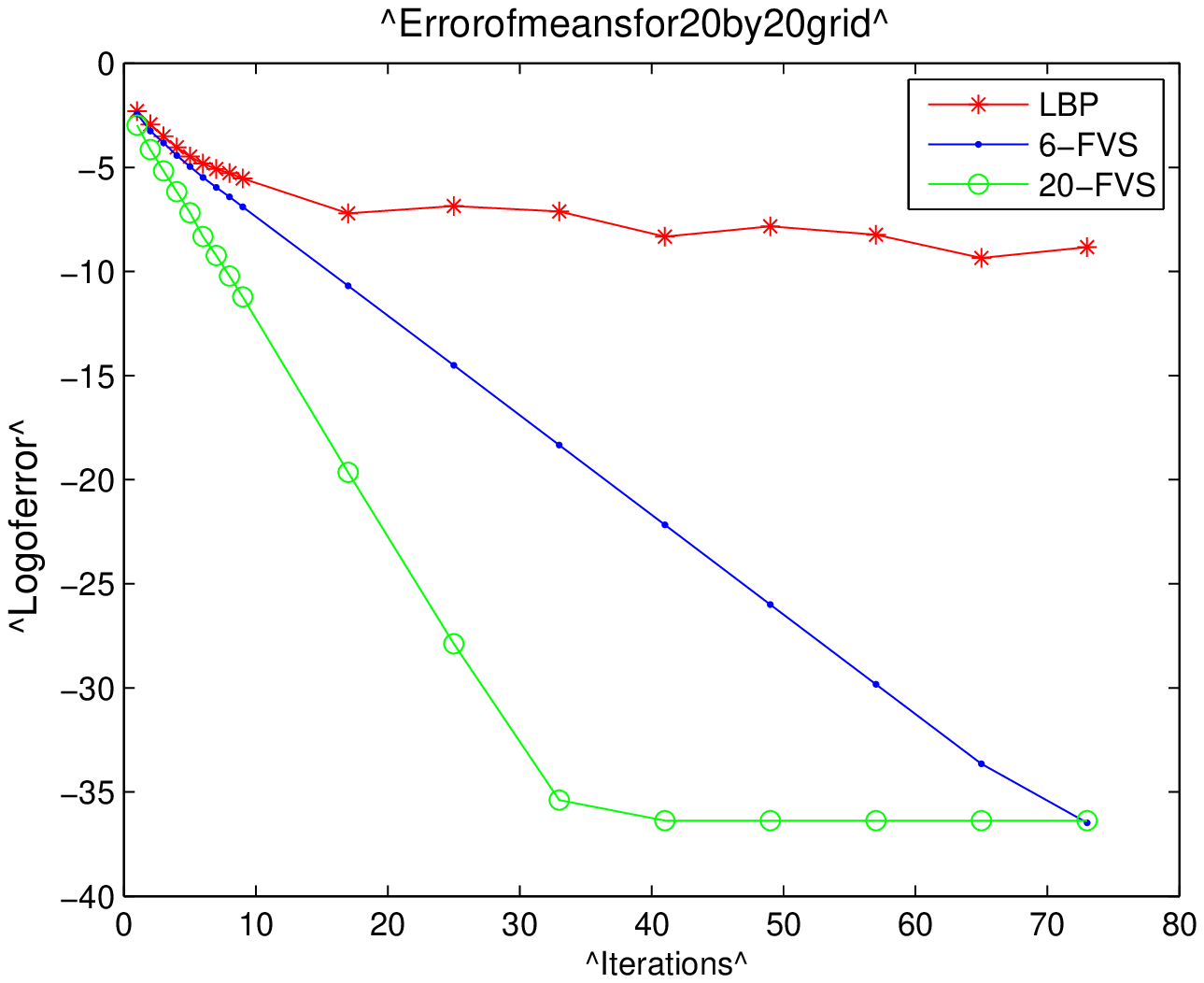}}
\caption{Inference errors of a $20\times 20$ grid}
\label{fig:good20}
\end{minipage}
\begin{minipage}{\columnwidth}
\centering
\vspace{15pt}
\subfigure[Evolution of variance errors with iterations]{
\label{fig:40-1}
\psfrag{^Iterations^}[c][c][1]{\small Iterations}
\psfrag{^Logoferror^}[c][c][1]{\small Log of error}
\psfrag{^Errorofvariancesfor40by40grid^}[c][c][1]{\small Error of means for $40\times 40$ grid}
\includegraphics[width=0.47\columnwidth]{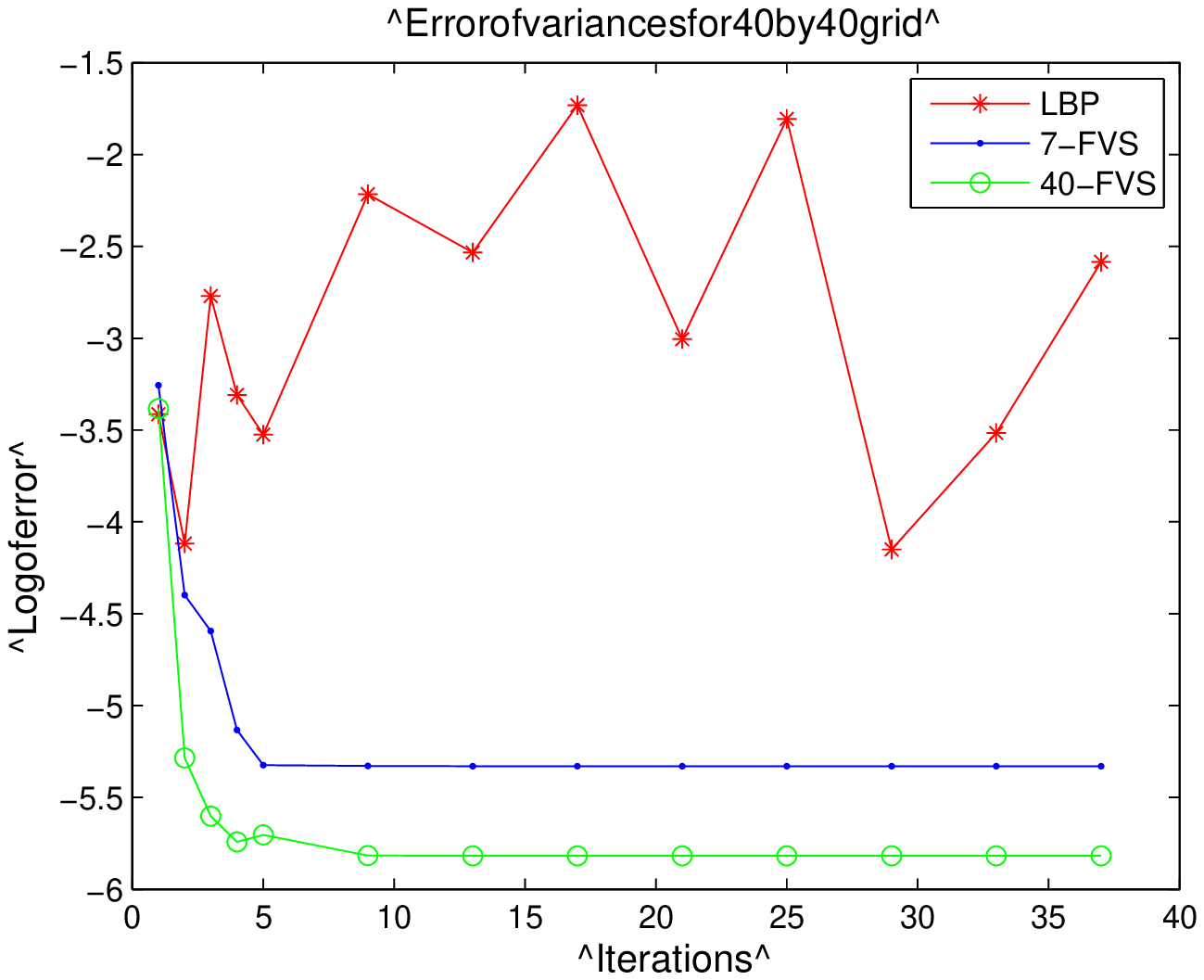}}
\subfigure[Evolution of mean errors with iterations]{
\label{fig:40-2}
\psfrag{^Iterations^}[c][c][1]{\small Iterations}
\psfrag{^Logoferror^}[c][c][1]{\small Log of error}
\psfrag{^Errorofmeansfor40by40grid^}[c][c][1]{\small Error of means for $40\times 40$ grid}
\includegraphics[width=0.47\columnwidth]{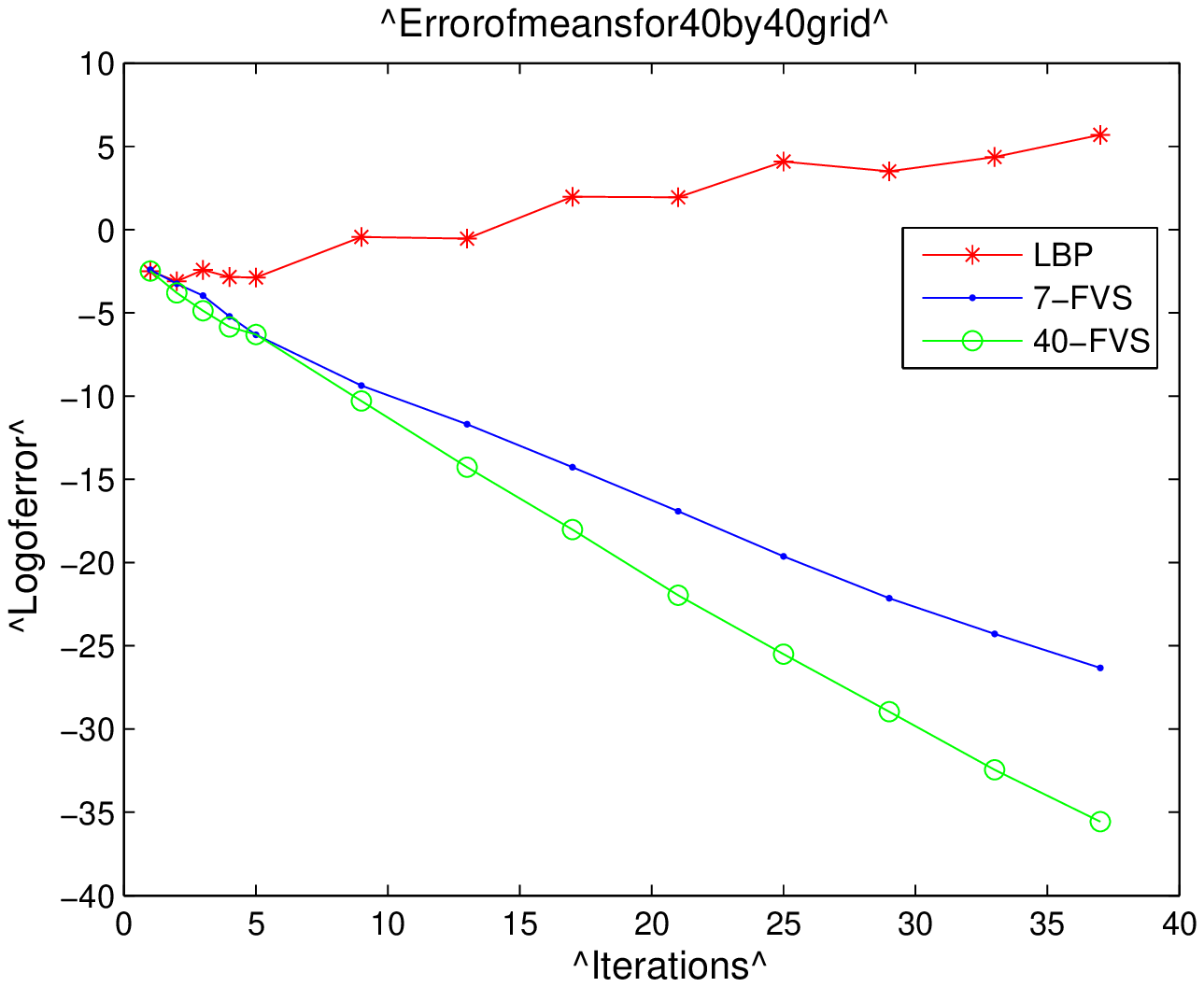}}
\caption{Inference errors of a $40\times 40$ grid}
\label{fig:good40}
\end{minipage}
\begin{minipage}{\columnwidth}
\centering
\vspace{15pt}
\subfigure[Evolution of variance errors with iterations]{
\label{fig:80-1}
\psfrag{^Iterations^}[c][c][1]{\small Iterations}
\psfrag{^Logoferror^}[c][c][1]{\small Log of error}
\psfrag{^Errorofvariancesfor80by80grid^}[c][c][1]{\small Error of means for $80\times 80$ grid}
\includegraphics[width=0.47\columnwidth]{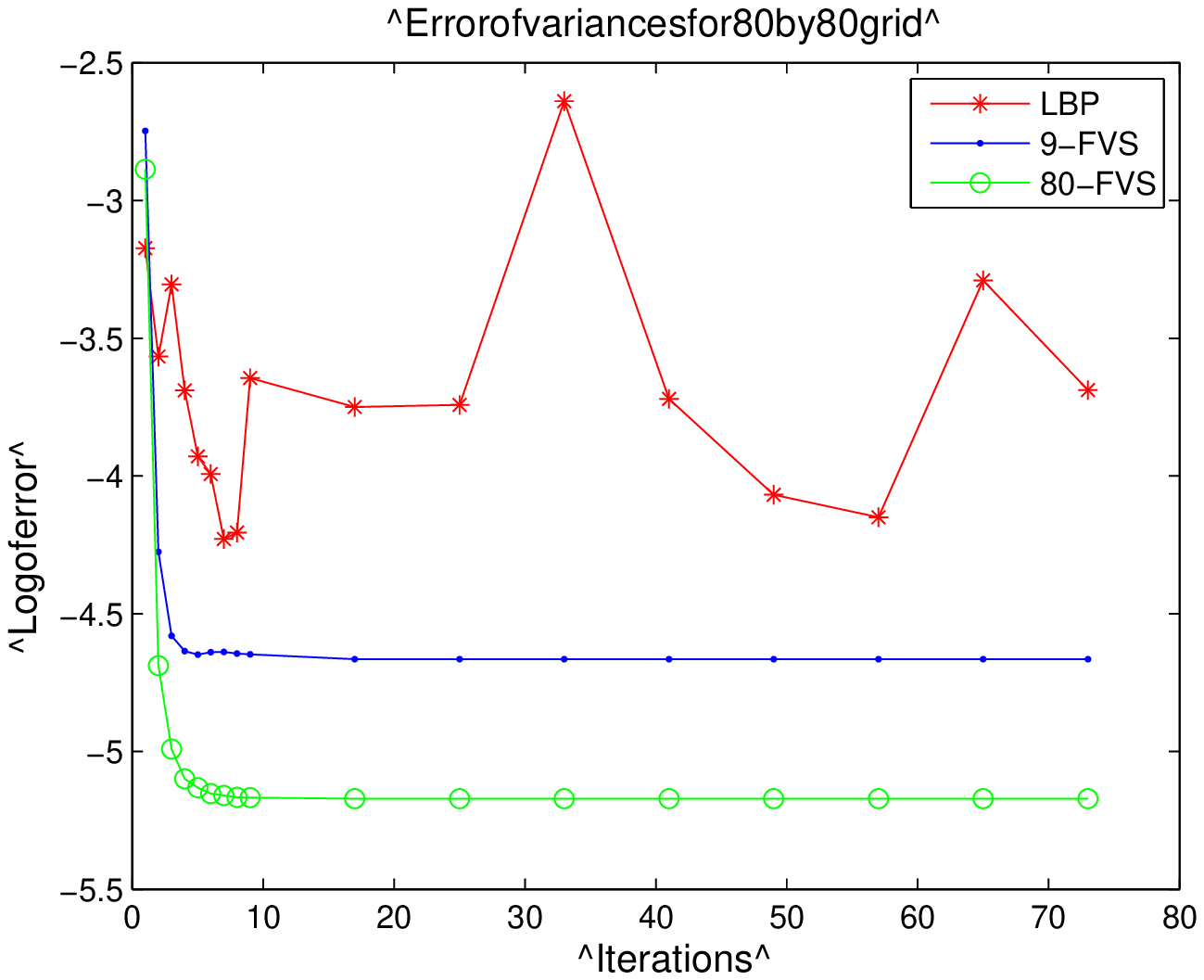}}
\subfigure[Evolution of mean errors with iterations]{
\label{fig:80-2}
\psfrag{^Iterations^}[c][c][1]{\small Iterations}
\psfrag{^Logoferror^}[c][c][1]{\small Log of error}
\psfrag{^Errorofmeansfor80by80grid^}[c][c][1]{\small Error of means for $80\times 80$ grid}
\includegraphics[width=0.47\columnwidth]{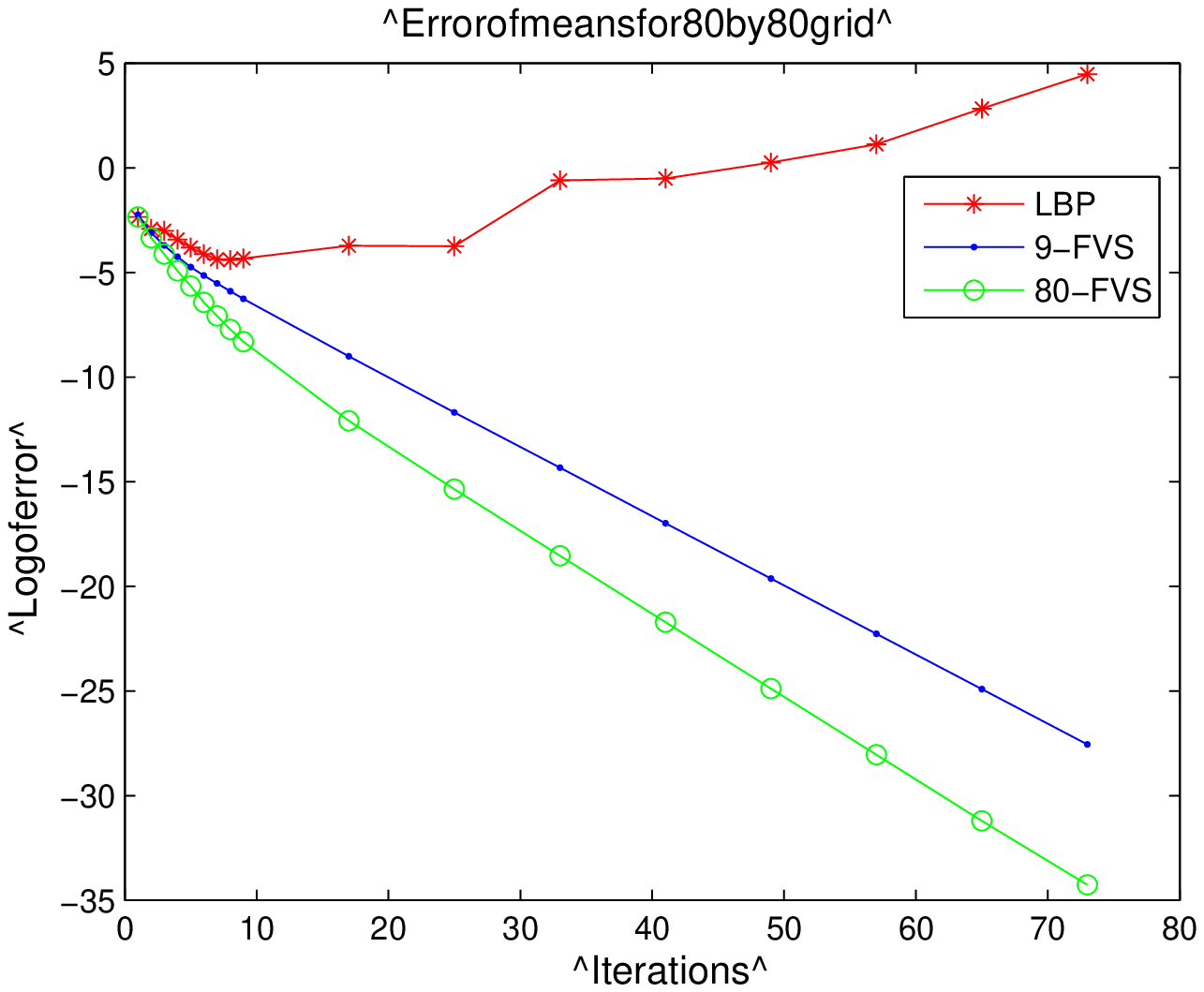}}
\caption{Inference errors of an $80\times 80$ grid}
\label{fig:good80}
\end{minipage}
\end{figure}

In Figures \ref{fig:good10a} to \ref{fig:good80}, numerical results are shown for $10\times 10$, $20\times 20$, $40\times 40$ and $80\times 80$ grids respectively.\footnote{Here we use shorthand terminology, where $k$-FVS refers to running our approximate FMP algorithm with a pseudo-FVS of cardinality $k$.} Except for the model in Figure \ref{fig:good10a}, LBP fails to converge for all models. With $k=\lceil \log n \rceil$ feedback nodes, approximate FMP converges for all the grids and gives much better accuracy than LBP. In Figure \ref{fig:good10a} where LBP converges on the original graph, we obtain more accurate variances and improved convergence rates using approximate FMP. In Figure \ref{fig:good10b} to \ref{fig:good80}, LBP diverges while approximate FMP gives inference results with small errors. When $k=\sqrt{n}$ feedback nodes are used, we obtain even better approximations but with more computations in each iteration. We performed approximate FMP on different graphs with different parameters, and empirically observed that $k=\lceil \log n \rceil$ feedback nodes seem to be sufficient to give a convergent algorithm and good approximations.  

\begin{figure}[t]
\begin{minipage}{1\columnwidth}
\centering
\subfigure[Evolution of variance errors with iterations]{
\label{fig:bad-1}
\psfrag{^Iterations^}[c][c][1]{\small Iterations}
\psfrag{^Logoferror^}[c][c][1]{\small Log of error}
\psfrag{^Errorofvariancesfor40by40grid^}[c][c][1]{\small Error of variances for $40\times 40$ grid}
\includegraphics[width=0.44\columnwidth]{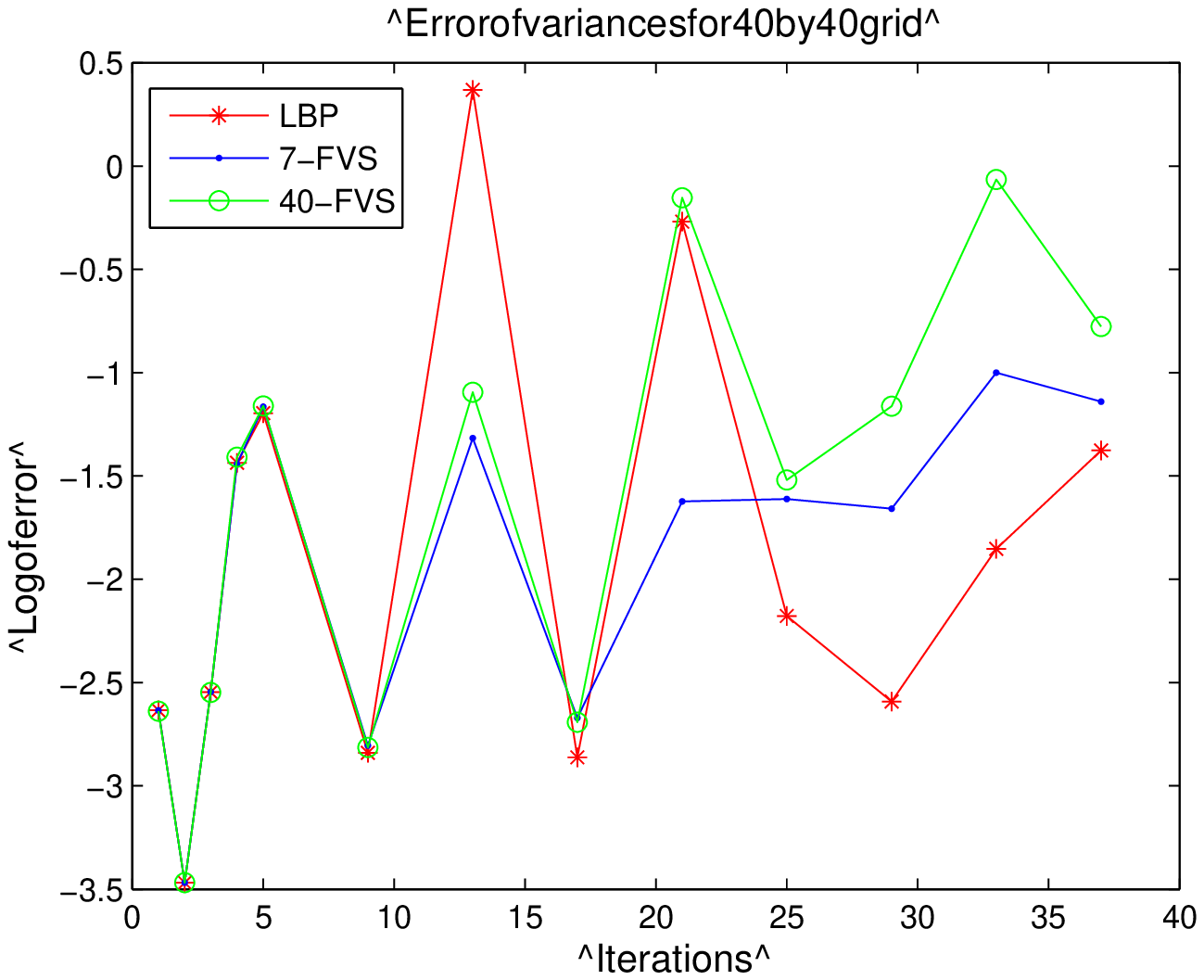}}
\subfigure[Evolution of means errors with iterations]{
\label{fig:bad-2}
\psfrag{^Iterations^}[c][c][1]{\small Iterations}
\psfrag{^Logoferror^}[c][c][1]{\small Log of error}
\psfrag{^Errorofmeansfor40by40grid^}[c][c][1]{\small Error of means for $40\times 40$ grid}
\includegraphics[width=0.44\columnwidth]{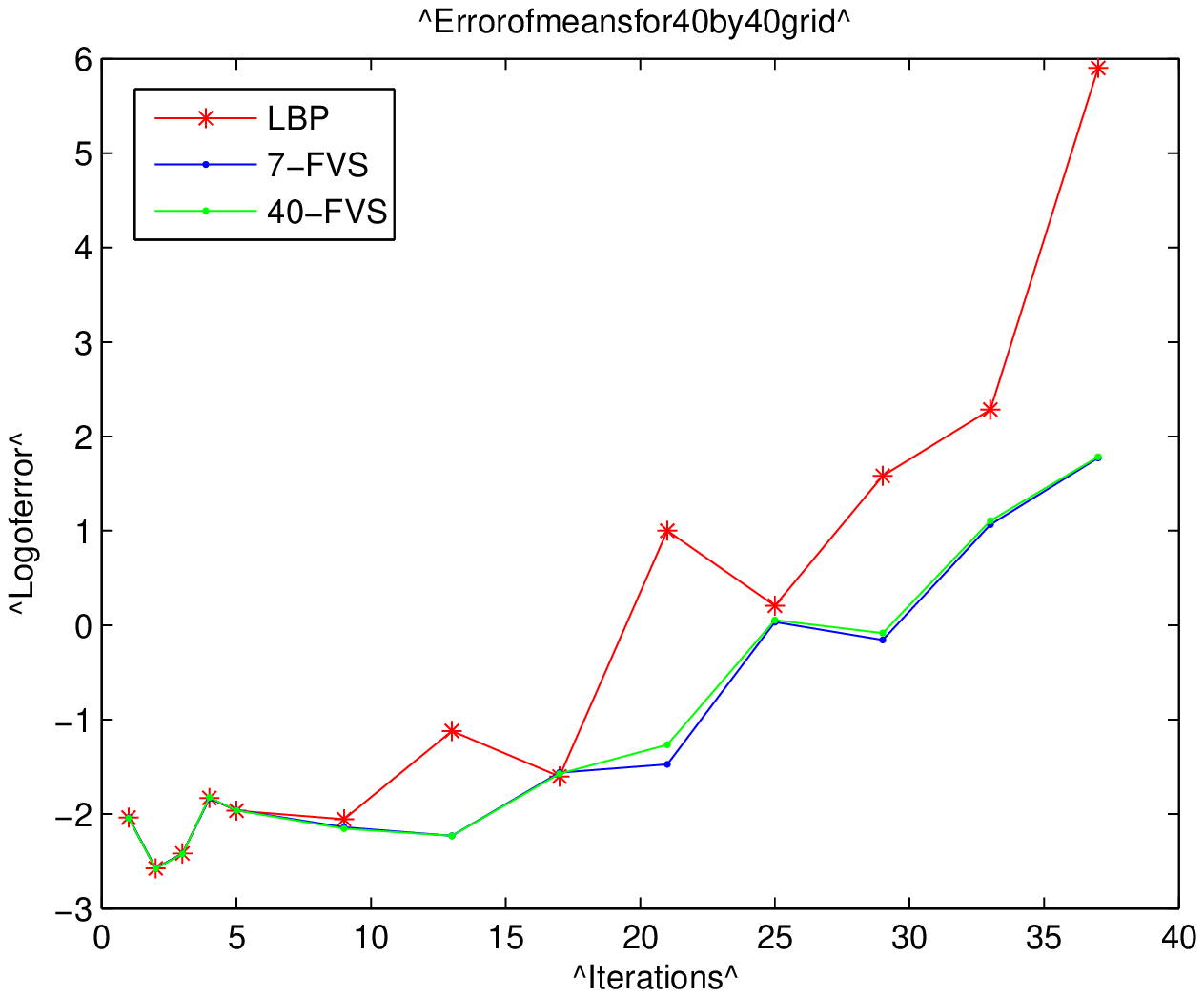}}
\caption{Inference errors with a bad selection of feedback nodes}
\label{fig:bad40}
\end{minipage}
\end{figure}

{\bf Remarks:} The question, of course, arises as to whether it is simply the {\em size} of the pseudo-FVS that is important. However, numerical results show that approximate FMP does not give satisfactory results if we choose a ``bad'' pseudo-FVS. In Figure \ref{fig:bad40}, we present results to demonstrate that the approximate FMP algorithm with a badly selected pseudo-FVS indeed performs poorly. The pseudo-FVS is selected by the opposite criterion of the algorithm in Figure \ref{fig:FVS1}, i.e., the node with the smallest score is selected at each iteration. We can see that LBP, 7-FVS, and 40-FVS algorithms all fail to converge. These results suggest that when a suitable set of feedback nodes are selected, we can leverage the graph structure and model parameters to dramatically improve the quality of inference in Gaussian graphical models.

\section{Conclusions and Future Directions}
\label{chap:conclusions}

In this paper we have developed the feedback message passing algorithm where we first identify a set of feedback nodes. The algorithm structure involves first employing BP algorithms on the remaining graph (excluding the FVS), although with several different sets of node potentials at nodes that are neighbors of the FVS; then using the results of these computations to perform exact inference on the FVS; and then employing BP on the remaining graph again in order to correct the answers on those nodes to yield exact answers. The feedback message passing algorithm solves the inference problem exactly in a Gaussian graphical model in linear time if the graph has a FVS of bounded size. Hence, for a graph with a large FVS, we propose an approximate feedback message passing algorithm that chooses a smaller ``pseudo-FVS'' and replaces BP on the remaining graph with its loopy counterpart LBP. We provide theoretical results that show that, assuming convergence of the LBP, we still obtain exact inference results (means and variances) on the pseudo-FVS, exact means on the entire graph, and approximate variances on the remaining nodes that have precise interpretations in terms of the additional ``walks'' that are collected as compared to LBP on the entire graph. We also provide bounds on accuracy, and these, together with an examination of the walk-summability condition, provide an algorithm for choosing nodes to include in the pseudo-FVS. Our experimental results demonstrate that these algorithms lead to excellent performance (including for models in which LBP diverges) with pseudo-FVS size that grows only logarithmically with graph size. 
 
There are many future research directions based on the ideas of this paper. For examples, more extensive study of the performance of approximate FMP on random graphs is of great interest. In addition, as we have pointed out, LBP is only one possibility for the inference algorithm used on the remaining graph after a pseudo-FVS is chosen. One intriguing possibility is to indeed use approximate FMP itself on this remaining graph -- i.e., nesting applications of this algorithm. This is currently under investigation, as are the use of these algorithmic constructs for other important problems, including the learning of graphical models with small FVS's and using an FVS or pseudo-FVS for efficient sampling of Gaussian graphical models.

\section{Acknowledgment}
We thank Devavrat Shah, Justin Dauwels, and Vincent Tan for helpful discussions.

\bibliographystyle{IEEEtran}
\bibliography{FVSJournal.bbl}
\end{document}